\pdfoutput=1
\documentclass[11pt]{article}

\usepackage{graphicx}
\usepackage{caption}
\usepackage{amsmath,amssymb}
\usepackage{booktabs}
\usepackage{multirow}
\usepackage{xcolor}
\usepackage{float}%
\usepackage[section]{placeins}%
\usepackage{pdfpages}
\usepackage[hidelinks]{hyperref}


\let\origappendix\appendix
\renewcommand{\appendix}{\origappendix
  \setcounter{figure}{0}%
  \renewcommand{\thefigure}{A.\arabic{figure}}%
  \renewcommand{\theHfigure}{appendix.A.\arabic{figure}}%
}

\title{Generative World Models Enable Predictive Control of Laser Melt Pool Dynamics}

\author{
Yiyang Yan$^{a}$,
Markus Bambach$^{a}$,
Mohamadreza Afrasiabi$^{a,*}$\\[1ex]
\textit{$^{a}$Advanced Manufacturing Lab, ETH Zürich, Leonhardstrasse 21, Zürich 8092, Switzerland}\\[1ex]
\textit{*~Corresponding author: afrasiabi@ethz.ch (M. Afrasiabi)}
}

\date{}

\begin{document}

\maketitle

\begin{abstract}
World models, which learn how environments respond to actions, are emerging as a powerful paradigm for planning through imagined futures, transforming decision-making across games, robotics and autonomous driving. Bringing this capability to manufacturing could enable process decisions on timescales inaccessible to high-fidelity simulation. Here we introduce a generative world model for localized highly dynamic laser melt pool that predicts evolution from histories of temperature and phase morphology under candidate actions. Its generative latent dynamics capture the effects of unresolved melt flow, enabling more accurate recursive rollouts than deterministic regressors under transient laser inputs. Because the learned dynamics are differentiable, the model can serve directly as a predictive control plant. Gradients through imagined futures optimize laser schedules that regulate melt-pool depth over previously unseen geometry, path, initialization. We further distil this optimization into an amortized policy that produces control actions in a single forward pass, providing a proof of concept for real deployment on machines.
\end{abstract}


\section{Introduction}
\label{sec:introduction}

Localized melt pools are the active transformation zones of fusion-based manufacturing, including welding, directed energy deposition, and laser powder bed fusion (LPBF) \cite{debroy_physical_1995,debroy_additive_2018}. In each process, concentrated energy input, fluid flow, phase change, and solidification interact within a transient region whose geometry and temperature directly influence thermal history, consolidation, and defect formation. LPBF is a particularly stringent case: micrometer-scale melt-pool dynamics evolving over microseconds interact with scan paths and parts spanning much larger spatial and temporal scales \cite{king_laser_2015,khairallah_laser_2016}, while small deviations can produce lack of fusion, keyhole porosity, or spatter \cite{huang_keyhole_2022,khairallah_controlling_2020}. High-fidelity thermal-fluid simulations resolve these mechanisms \cite{muther_identifying_2025,zhang2026mixing}, but even GPU-accelerated solvers need minutes to hours per simulation \cite{luthi2025multi3,leonor2024gomelt}, far beyond the timescale of control decisions.

Controlling the three-dimensional melt-pool geometry, rather than surface signals, would directly target lack of fusion and overmelting, but is currently infeasible from both measurement and simulation. In-process sensors capture only surface signals, and operando X-ray imaging, although able to reconstruct four-dimensional melt pools at 25~kHz, requires a synchrotron beamline and offline reconstruction \cite{witte_4d_2026}. In simulation, a linear surrogate identified from smoothed particle hydrodynamics enabled depth control in two dimensions, but its extension to three dimensions was computationally prohibitive \cite{liao-mcpherson_layer--layer_2025}. Controllers therefore act at the layer \cite{KAVAS2026105124} or vector \cite{kirschbaum2025vector} level or on conduction models \cite{ogoke_thermal_2021,mozaffar_differentiable_2023}, and neural surrogates predict steady-state \cite{benoit_fast_2026} or top-view \cite{manav_meltpoolinr_2025} morphology, or three-dimensional dynamics without actions \cite{zhang_scientific_2026}. None combines recursive 3D rollout, time-varying actions, partial observations and usable action gradients.

World Models (WM) enable this combination. They learn a compact state and an action-conditioned transition, so that behaviour can be optimized in imagined trajectories \cite{ha_recurrent_2018}, by online planning \cite{hansen_temporal_2022} or by gradients through the learned dynamics \cite{hafner_dream_2020}. Under partial observation, reliable planning required both a deterministic memory of past observations and a stochastic transition \cite{hafner_learning_2019}, and diffusion transitions now preserve details important for control \cite{alonso_diffusion_2024} and simulate games in real time \cite{valevski_diffusion_2025}. The melt pool fits this setting: our model resolves temperature and phase morphology but not the velocity and pressure fields driving convective and Marangoni transport, so identical observed states can evolve differently. Eliminating such hidden variables yields reduced dynamics with memory and fluctuating terms \cite{mori_transport_1965,zwanzig_nonlinear_1973}. A latent history can capture the memory, and a generative transition the range of futures consistent with it.

When a data-driven model is used in place of the plant for control, it can fail in two ways. Deterministic surrogates trained with mean-squared error average over hidden futures, regressing toward the mean flow in turbulence \cite{molinaro2025generativeaifastaccurate} and losing fine-scale content needed for accurate long rollouts \cite{lippe_pde-refiner_2023}. Generative models avoid this averaging but can instead produce plausible states that violate the underlying physics \cite{kang_how_2025}. Control amplifies both errors because an optimizer seeks actions that perform well under the learned model regardless of whether its predictions are correct, while policies trained in learned dynamics can exploit poorly represented regions \cite{kurutach_model-ensemble_2018}. Whether an action-conditioned generative transition provides gradients reliable enough for control, and whether the resulting actions remain effective when re-evaluated by the physics solver, remains largely unexamined.

Here we introduce a WM with generative latent dynamics for transient 3D melt-pool prediction and control. It encodes temperature and phase-indicator fields into a compact finite-scalar-quantized (FSQ) latent state \cite{mentzer_finite_2023,lingsch_phaedra_2026}, which is advanced by a history-conditioned flow-matching transition differentiable with respect to the laser actions. We validate that the differentiable latent dynamics remain accurate over control-relevant rollouts and provide useful gradients for model predictive control, and further show that an RL policy can reach the same objective in a single forward pass to realize faster planning. The control target is melt-pool depth over variable substrate geometries, scanning path and residual heat, where insufficient or excessive depth causes severe defects (e.g lack of fusion, overmelting). The selected schedules are re-evaluated in high-fidelity CFD to test whether control performance persists outside the learned model.

The melt-pool WM presents a technical proof-of-concept that localized, highly dynamic and partially observed melt pools can be learned well enough to support gradient-based control, as verified in high-fidelity simulation. We also discuss potential use-cases, including within-track in-line control on LPBF machines and other fusion-based processes such as welding and directed energy deposition. However, operationalizing the WM for these applications will require further innovation, notably a measurement-conditioned encoder that maps in-process sensor signals to the latent state and inference fast enough for the melt-pool timescale, in addition to open machine platforms with synchronized sensing and laser actuation.
\section{Results}
\label{sec:results}

\subsection{Generative latent world model for partial thermal-fluid states}
\label{sec:res_overview}
\begin{figure}[!htb]
  \centering
  \includegraphics[width=\linewidth]{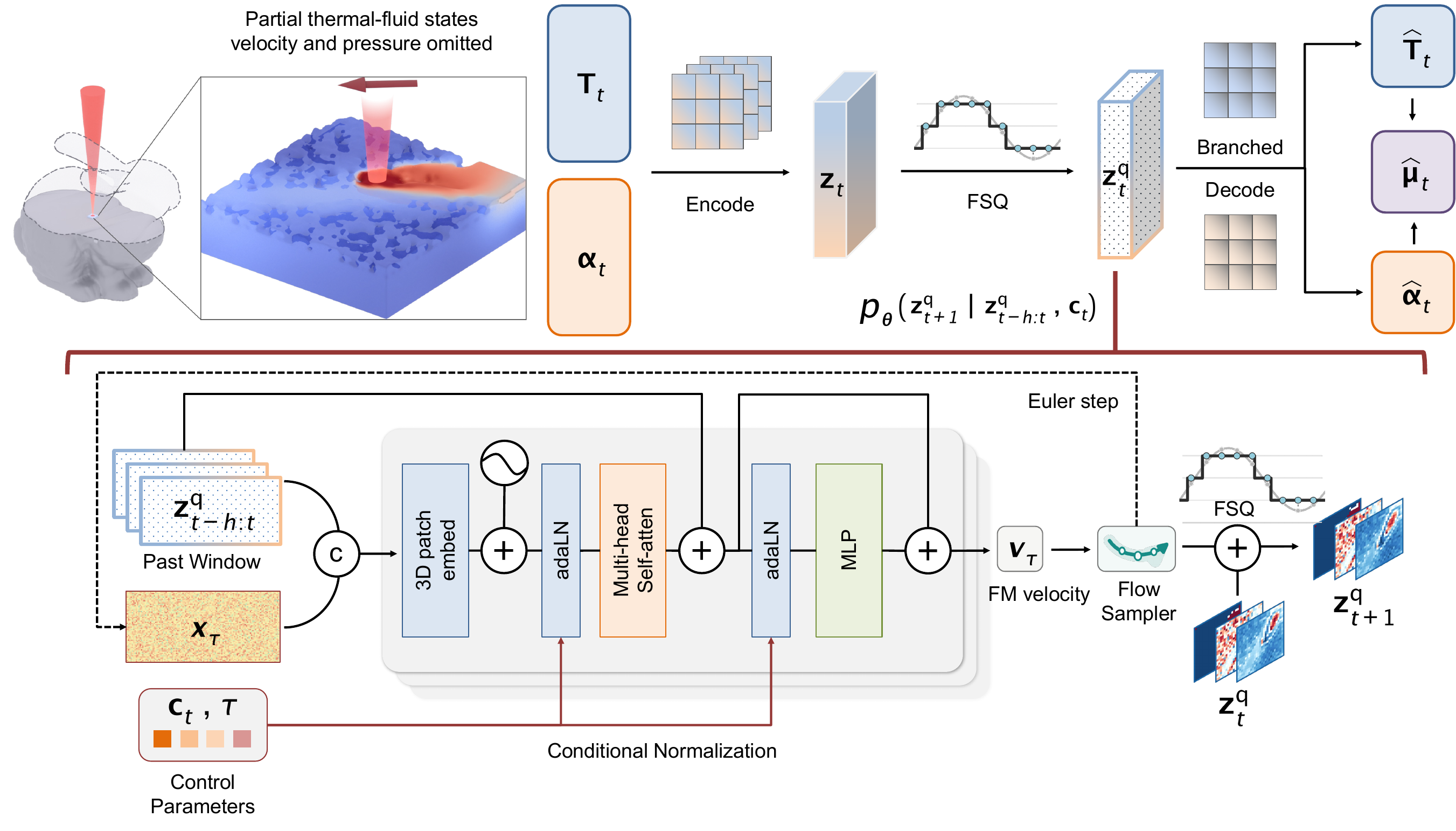}
  \caption{\textbf{Melt-pool WM with generative latent dynamics.} \textbf{Top:} the autoencoder compresses the partially observed state $\mathbf{X}_t=(\mathbf{T}_t,\boldsymbol{\alpha}_t)$, without the full CFD thermal-fluid state, into a discrete code $\mathbf{z}^{\mathrm{q}}_t$. Two decoder branches reconstruct the smooth $\hat{\mathbf{T}}_t$ and the sharp $\hat{\boldsymbol{\alpha}}_t$ from the shared latent code, and the liquid fraction $\hat{\boldsymbol{\mu}}_t$ is computed from the reconstructed fields. \textbf{Bottom:} the transition $p_{\theta}(\mathbf{z}^{\mathrm{q}}_{t+1}\mid \mathbf{z}^{\mathrm{q}}_{t-h:t},\mathbf{c}_t)$: diffusion-transformer (DiT) blocks, conditioned on the control parameters $\mathbf{c}_t$ and scalar flow-matching time $\tau$ through adaptive layer normalization (adaLN), regress the flow-matching velocity $\mathbf{v}_{\tau}$, which an Euler sampler integrates into a latent residual that is added to $\mathbf{z}^{\mathrm{q}}_t$ and re-quantized.}
  \label{fig:wm_details}
\end{figure}

The WM evolves the transient three-dimensional melt pool in a learned latent space (Fig.~\ref{fig:wm_details}). An encoder maps the temperature and phase-indicator fields $(\mathbf{T},\boldsymbol{\alpha})$ to a FSQ latent grid, in which each channel is rounded to a finite set of levels, so that one compact code accommodates both the sharp phase interfaces and the smoother temperature field. A history-conditioned flow-matching transition, a generative model that learns to transform Gaussian noise into samples of the next latent state, advances this code under the laser actions and accounts for the influence of the unobserved velocity and pressure fields. The decoder reconstructs physical fields only when quantities such as temperature, phase morphology, or melt-pool depth are evaluated (Methods).

\subsection{Rollout fidelity and generalization across different regimes}
\label{sec:res_surrogate}

\begin{figure}[p]
  \centering
  \includegraphics[width=0.92\textwidth]{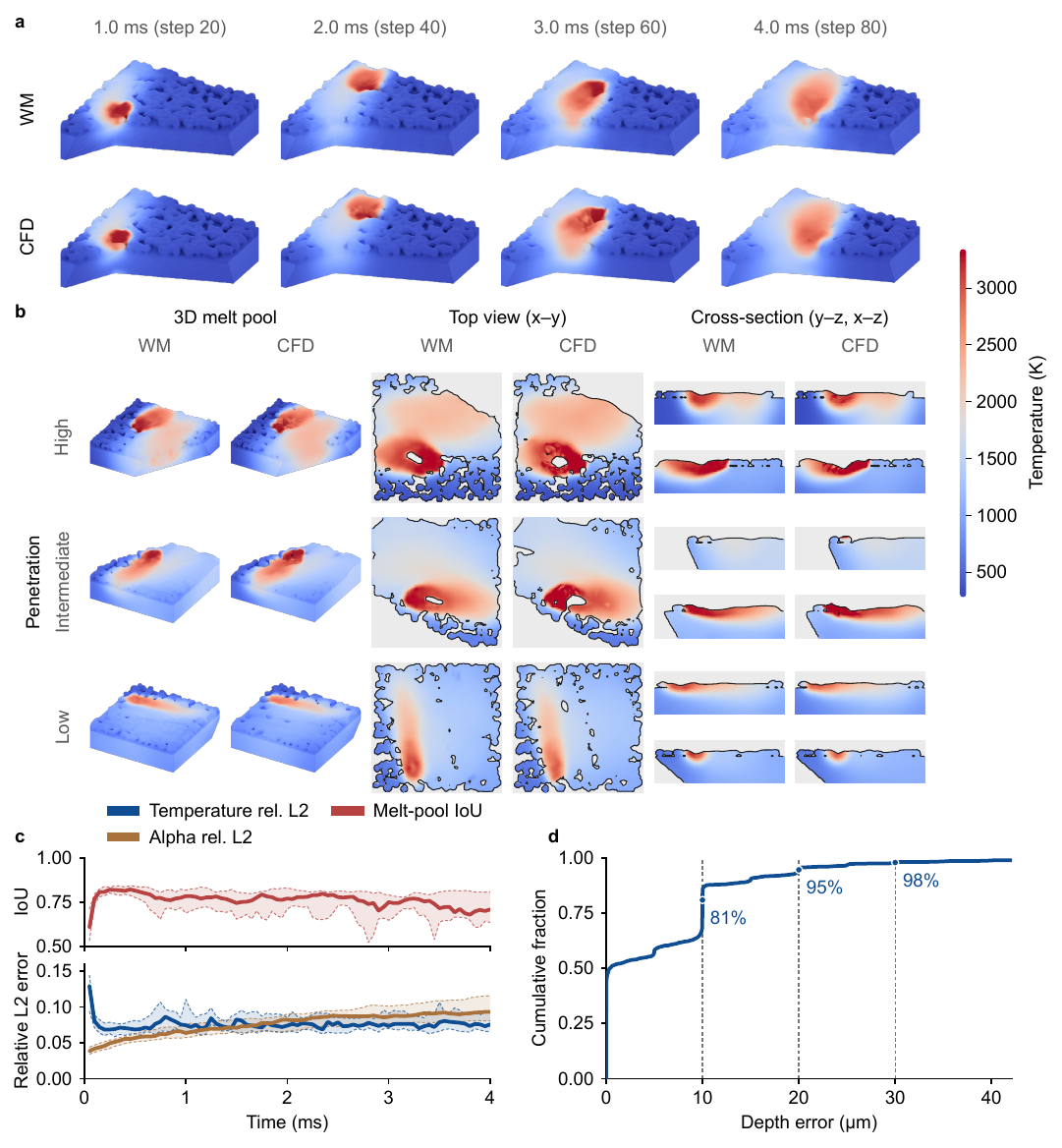}
  \caption{\textbf{World-model fidelity, stability and generalization.} All panels show the base WM, pretrained on constant schedules, on the constant-parameter test cases. \textbf{(a)} Recursive rollout on a representative scan: WM versus CFD temperature, rendered on the phase indicator $\boldsymbol{\alpha}$ (material versus air), at four rollout steps. \textbf{(b)} Final-frame WM versus CFD for three cases selected to span high, intermediate and low final penetration, shown as the three-dimensional melt pool, top view ($x$-$y$) and depth cross-sections ($y$-$z$ and $x$-$z$). The computational domain spans $1~\mathrm{mm}\times1~\mathrm{mm}\times0.32~\mathrm{mm}$. \textbf{(c)} Temperature and phase-indicator relative $L_2$ error and melt-pool intersection over union versus rollout time, aggregated across cases. Lines show the mean across cases and bands show the interquartile range across cases. \textbf{(d)} Cumulative distribution of melt-pool depth error over all cases, with dotted lines marking one, two and three voxels ($\mathrm{d}x=\mathrm{d}y=\mathrm{d}z=10~\mu\mathrm{m}$).}
  \label{fig:res_surrogate}
\end{figure}

We first test whether the latent dynamics remain accurate under recursive prediction, where errors can accumulate over many steps (Fig.~\ref{fig:res_surrogate}). In a representative held-out trajectory, the WM follows the moving thermal field and reproduces the location and extent of the high-temperature region from step 20 to step 80 without visible drift (Fig.~\ref{fig:res_surrogate}a).

The model generalizes to unseen substrate geometries, scan paths, and initial temperature fields (Fig.~\ref{fig:extended_initial_conditions}). Across high, intermediate, and low final penetration, it recovers the extended melt region penetrating the substrate, the smaller localized melt pool, and the shallow melt pool without spurious depth, in the three-dimensional rendering as well as in the top views and cross-sections (Fig.~\ref{fig:res_surrogate}b).

Aggregate metrics confirm this stability (Fig.~\ref{fig:res_surrogate}c). Temperature error remains bounded over the horizon, whereas the phase-indicator error roughly doubles over $4~\mathrm{ms}$ as small interface discrepancies accumulate. Melt-pool intersection over union (IoU) stays between about $0.7$ and $0.8$, and its larger inter-case spread reflects the sensitivity of the phase boundary to small local errors near the solid--liquid interface. Under transient inputs, errors grow steadily over the horizon for all models (Fig.~\ref{fig:res_stepper}c-e,h-j).

For control, the decisive quantity is whether the model predicts penetration accurately enough to distinguish under- from over-penetration. Pooling per-step predictions from $25$ held-out rollouts of $80$ steps ($n=2000$), with depth taken as the median over an eight-member ensemble, more than $80\%$ of predicted depths lie within one voxel ($10~\mu\mathrm{m}$) of CFD, $95\%$ within two voxels, and $98\%$ within three (Fig.~\ref{fig:res_surrogate}d). Because the decoded liquid fraction changes sharply across the melt-pool boundary, depth errors cluster near multiples of the vertical grid spacing, so the effective depth resolution is about one grid cell, a fifth of the $50~\mu\mathrm{m}$ control target. The learned dynamics therefore remain stable under recursion, generalize across penetration regimes, and preserve melt-pool geometry at an accuracy suitable for depth control.

\subsection{Discrete tokenization, generative latent dynamics and adaptation to transient melt pool}
\label{sec:res_ablation}

\begin{figure}[!htbp]
  \centering
  \includegraphics[width=0.95\textwidth]{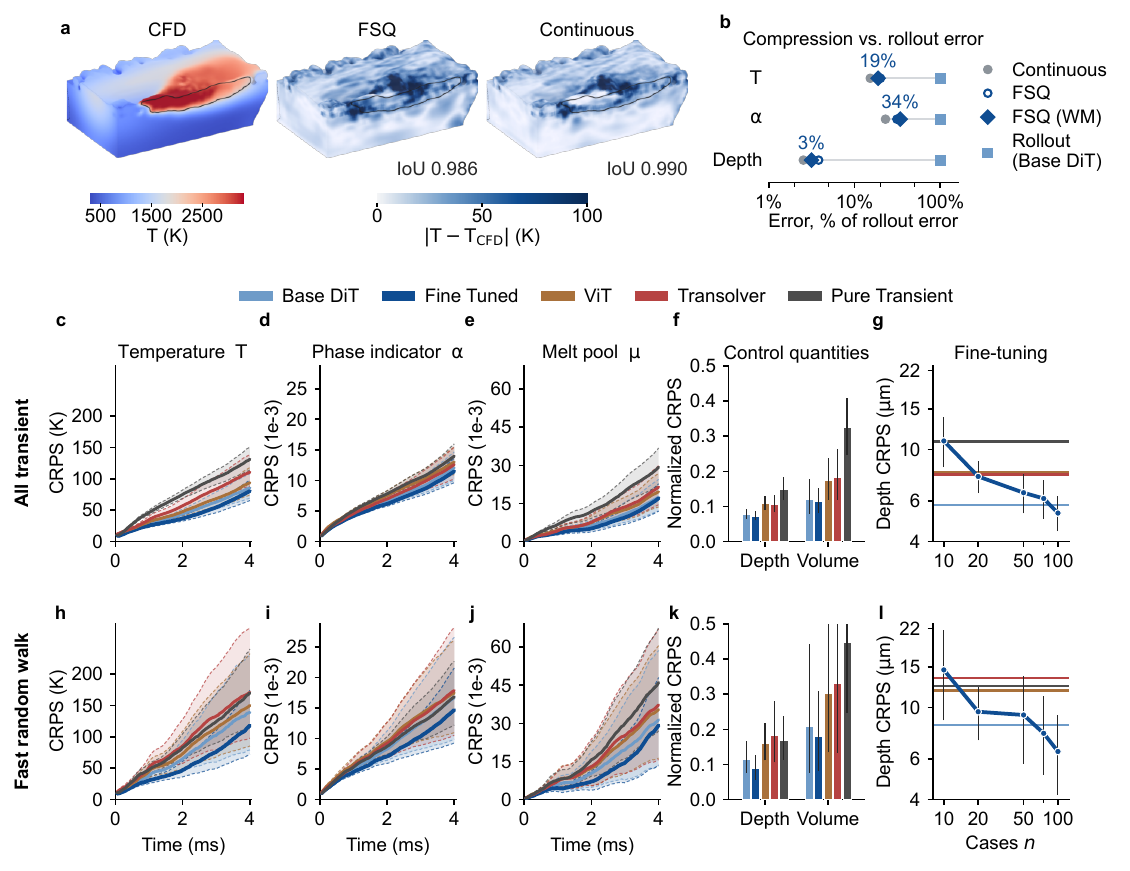}
  \captionsetup{font=small,skip=6pt}
  \caption{\textbf{A compact discrete representation at small reconstruction cost, and generative dynamics that support prediction under transient laser inputs.} \textbf{(a)} Reconstruction of a held-out constant-parameter frame ($t=1.7~\mathrm{ms}$), chosen at the median melt-pool-field error of the FSQ autoencoder: CFD temperature with the melt-pool boundary ($\boldsymbol{\alpha}\boldsymbol{\mu}=0.5$) outlined, and absolute temperature error $|\mathbf{T}-\mathbf{T}_{\mathrm{CFD}}|$ of the FSQ autoencoder and of a continuous bottleneck of identical architecture. IoU, melt-pool intersection over union with CFD. \textbf{(b)} Reconstruction error of the continuous autoencoder, the FSQ autoencoder for ablation and the FSQ autoencoder of the WM (WM), relative to the rollout error of the base WM (100\%), on the $88$ transient test cases: median relative $L_2$ error of temperature and phase indicator, and melt-pool depth error (depth CRPS for the rollout). FSQ denotes the pretrained autoencoder and FSQ (WM) the same autoencoder fine-tuned on time-varying frames and used by all transition models (Supplementary Tables~S5 and~S8). \textbf{(c-g)} Transition-model evaluation on all $88$ transient test cases. \textbf{(h-l)} As \textbf{c-g}, on the $19$ fast random-walk cases. \textbf{(c,h)} Temperature, \textbf{(d,i)} phase-indicator and \textbf{(e,j)} melt-pool-field continuous ranked probability score (CRPS) over the $4~\mathrm{ms}$ autoregressive rollout. \textbf{(f,k)} Melt-pool depth and volume CRPS pooled over all laser-on steps, normalized in both rows by the mean CFD value over the laser-on steps of all $88$ transient cases. \textbf{(g,l)} Laser-on depth CRPS, pooled as in \textbf{f,k}, after fine-tuning on $n=10$, $20$, $50$, $75$ or $100$ time-varying cases. Horizontal lines show the reference models. Base DiT is the flow-matching transition pretrained on constant schedules ($360$ training and $40$ validation cases), and Fine Tuned is its early-stopped adaptation to $100$ time-varying cases, used by both controllers. ViT \cite{dosovitskiy_image_2021} and Transolver \cite{wu_transolver_2024} are deterministic baselines trained on the same constant-schedule data. Pure Transient is the flow-matching transition trained from scratch on all $800$ time-varying cases, including the fine-tuning and validation cases. Generative models use eight rollout samples per case, whereas for deterministic models CRPS reduces to the mean absolute error. Curves in \textbf{c-e} and \textbf{h-j} show case means, and bands and error bars in \textbf{c-l} indicate $95\%$ bootstrap confidence intervals over cases.}
  \label{fig:res_fsq}
  \label{fig:res_stepper}
\end{figure}

Discretizing the latent space typically costs reconstruction fidelity \cite{rombach_high-resolution_2022,li_autoregressive_2024}, but discrete latents have benefited learned WMs in downstream task performance \cite{hafner2021mastering} and in multi-step prediction when model capacity is limited \cite{meyer_harnessing_2024}. Against a continuous bottleneck of identical architecture trained on the same constant-schedule frames, the pretrained FSQ autoencoder raises the median reconstruction error on unseen cases by $0.4$, $0.6$-$0.7$ and $1.6$-$1.8$ percentage points for temperature, phase indicator and melt-pool field (Supplementary Table~S8). In a representative frame, both autoencoders recover the melt pool with an IoU of $0.99$ and differ mainly in small temperature errors around it, with medians of $5$ and $4~\mathrm{K}$ in the metal (Fig.~\ref{fig:res_fsq}a). This cost is small relative to the dynamics. Encoding and decoding the CFD fields with the fine-tuned world-model autoencoder gives a depth error of $0.17$-$0.18~\mu\mathrm{m}$ and a temperature relative $L_2$ error of $1.7\%$, about $30$ and $5$ times below the rollout's depth continuous ranked probability score (CRPS, $4.5$-$5.8~\mu\mathrm{m}$) and temperature error ($7.5$-$9.0\%$). The CRPS measures how far a predicted distribution lies from the CFD value and reduces to the mean absolute error for a deterministic prediction. With a phase-indicator error about one third of the rollout's, the representation is not the accuracy bottleneck of the WM (Fig.~\ref{fig:res_fsq}b).

The transition comparison tests how the models handle the hidden flow. The deterministic ViT \cite{dosovitskiy_image_2021} and Transolver \cite{wu_transolver_2024} regressors reach lower one-step validation errors than the flow-matching model (Supplementary Fig.~S4a), as expected for predictors of the conditional mean, yet over recursive rollouts on the transient test cases the flow-matching model is more accurate in temperature and melt-pool field, and the gap grows with the horizon (Fig.~\ref{fig:res_stepper}c,e). This is consistent with conditional-mean predictions leaving the manifold of realizable states and compounding under recursion, whereas the generative transition samples a realizable state at each step. The slowly evolving phase indicator, largely determined by the observed history, is predicted similarly by all models (Fig.~\ref{fig:res_stepper}d). The advantage carries over to both control quantities: the normalized CRPS of the flow-matching model is $26$-$28\%$ lower for melt-pool depth and approximately $33\%$ lower for melt-pool volume than that of the two deterministic baselines (Fig.~\ref{fig:res_stepper}f). Paired over the $88$ transient cases, the depth CRPS is $2.2$ and $2.0~\mu\mathrm{m}$ lower than that of the ViT and the Transolver (both $P<10^{-3}$), and the volume CRPS is $0.48\times10^{-3}$ and $0.58\times10^{-3}~\mathrm{mm^{3}}$ lower ($P=2\times10^{-4}$ and $P<10^{-3}$). On the fast random walk (Fig.~\ref{fig:res_stepper}h-k), the depth gap widens to $3.4~\mu\mathrm{m}$ ($P=0.046$) and $5.0~\mu\mathrm{m}$ ($P<10^{-3}$).

Pretrained on constant schedules, the WM already generalizes to step and ramp inputs, between which the melt pool relaxes toward a quasi-steady state (Fig.~\ref{fig:res_stepper}c-f). A controller that updates power and speed at every interval, however, induces inputs and states closer to the fast random walk, where the melt pool rarely reaches a steady state and every model shows its largest and fastest-growing errors (Fig.~\ref{fig:res_stepper}h-l). Fine-tuning on time-varying schedules targets this regime but requires sufficient data: $n=10$ cases roughly double the depth error, and only with $n=100$ cases is the fine-tuned model more accurate than the base model (Fig.~\ref{fig:res_stepper}g,l). The gain concentrates where control needs it. On the fast random walk, the depth CRPS falls from $8.4$ to $6.5~\mu\mathrm{m}$, the largest change among the four regimes, although not significant with $19$ cases ($P=0.29$), and the temperature and melt-pool-field errors separate from those of the base model after approximately $2~\mathrm{ms}$. Step and ramp inputs do not benefit (Supplementary Fig.~S5). Qualitative fast-random-walk rollouts, including a case where fine-tuning worsens depth accuracy, are shown in Supplementary Figs.~S7--S10. The adaptation costs little on constant-parameter inputs: the fraction of depth predictions within one voxel of CFD decreases from $81\%$ to $74\%$, and within $30~\mu\mathrm{m}$ from $98\%$ to $96\%$ (Supplementary Fig.~S6).

Exposure to time-varying inputs alone does not ensure this accuracy. Trained from scratch on all $800$ time-varying cases of the pool from which the fine-tuning cases are drawn, including the fine-tuning and validation cases, the Pure Transient model has the highest melt-pool volume CRPS on the fast random walk ($4.11\times10^{-3}~\mathrm{mm^{3}}$, compared with $1.92\times10^{-3}$ for the base and $1.65\times10^{-3}~\mathrm{mm^{3}}$ for the fine-tuned model), a higher depth CRPS ($12.4$ versus $8.4$ and $6.5~\mu\mathrm{m}$), and larger melt-pool-field errors than either pretrained model (Fig.~\ref{fig:res_stepper}j-l). One possible explanation is that constant-schedule pretraining first establishes melt-pool formation and relaxation toward quasi-steady states, on which the perturbations caused by changing inputs are easier to learn under partial observation than when rapid forcing and unobserved thermal-fluid relaxation must be learned together. Because a controller issues time-varying inputs, we use the fine-tuned model for both controllers.

\subsection{Control optimization and CFD re-validation}
\label{sec:res_control}
\begin{figure}[!tbp]
  \centering
  \includegraphics[width=\textwidth]{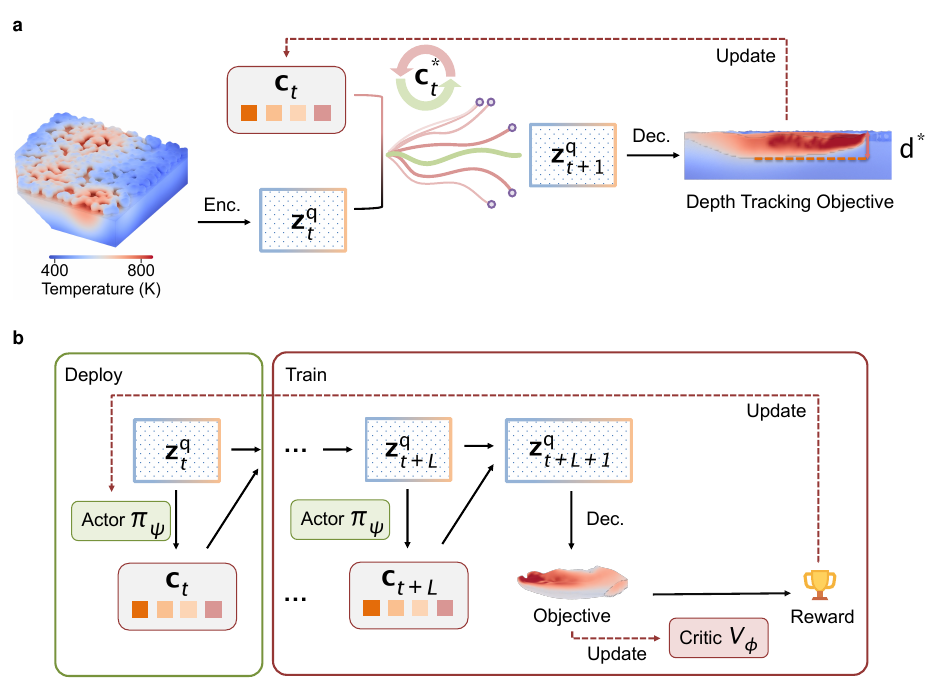}
  \caption{\textbf{Two controllers operating on the same latent WM.} \textbf{(a)} At each control step, the gradient-based receding-horizon controller encodes the current observation into the latent state $\mathbf{z}^{\mathrm{q}}_t$ and rolls the frozen WM forward under a candidate control sequence $\mathbf{C}_t$. The predicted latent state is decoded to evaluate the melt-pool depth tracking objective against the target $d^*$, and gradients are propagated through the rollout to update the control sequence. \textbf{(b)} The latent policy is trained within the same frozen WM using SHAC. The actor $\pi_\psi$ generates power and speed commands from the latent state, while imagined rollouts provide the objective and reward used to update the actor and critic $V_\phi$. After training, only the actor is required for deployment, mapping the current latent state directly to the control command in a single forward pass.}
  \label{fig:controllers}
\end{figure}

We next test whether the differentiable WM can select laser power and scan speed that remain effective when transferred back to CFD. Both controllers use the same frozen WM, the fine-tuned transition with $n=100$ cases of Fig.~\ref{fig:res_stepper}, but in different ways (Fig.~\ref{fig:controllers}). The receding-horizon controller (MPC) rolls the latent dynamics forward under a candidate control sequence and backpropagates the depth-tracking objective through the rollout. The amortized policy (RL) is trained with short-horizon actor-critic (SHAC) \cite{xu2021accelerated}, which backpropagates through short imagined rollouts, and maps the current latent state directly to a control command. MPC thus solves the control problem repeatedly during deployment, whereas the policy learns it in advance and executes it in a single forward pass.

\begin{figure}[!tbp]
  \centering
  \includegraphics[width=\textwidth]{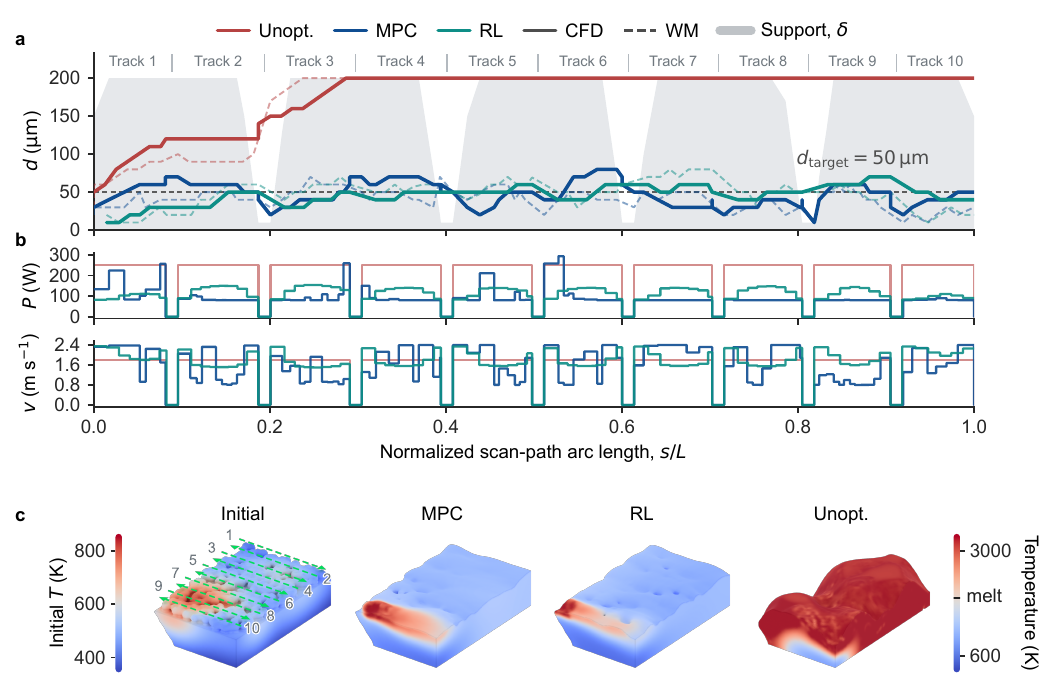}
  \caption{\textbf{World-model control and CFD re-validation in a representative overhang case.} \textbf{(a)} Melt-pool depth along the normalized scan path for the unoptimized scan, gradient receding-horizon controller (MPC), and amortized latent policy (RL), together with the local support thickness $\delta$ and target depth $d_{\mathrm{target}}=50~\mu\mathrm{m}$. Solid curves show CFD re-simulations and dashed curves the corresponding world-model rollouts. The unoptimized scan progresses toward full penetration, whereas both controllers maintain the melt-pool depth near the target across the ten tracks. \textbf{(b)} Laser-power and scan-speed schedules selected by MPC and RL, with the unoptimized settings shown for reference. \textbf{(c)} Three-dimensional temperature fields for the common initial state and the final CFD states obtained with MPC, RL, and the unoptimized scan. Both controlled cases suppress the large high-temperature region observed without control.}
  \label{fig:res_control}
\end{figure}

The stringent test is whether decisions optimized through the learned dynamics retain their effect in CFD (Fig.~\ref{fig:res_control}). In a representative ten-track overhang case, the unoptimized scan, which holds the nominal $P=250~\mathrm{W}$ and $v=1.8~\mathrm{m\,s^{-1}}$ that also initialize both controllers, reaches full penetration after the first few tracks and remains there for much of the scan. It serves as a reference for the thermal disturbance accumulating along the overhang rather than as a baseline for the learned dynamics. Both MPC and RL instead regulate the depth around the $50~\mu\mathrm{m}$ target despite changes in local support thickness between tracks, and the world-model rollouts follow the CFD re-simulations with local deviations. Because these trajectories are generated by actions that the model itself selected, the comparison directly tests the learned dynamics under control-induced state and action distributions.

The controllers reach this regulation through different schedules (Fig.~\ref{fig:res_control}b): MPC makes abrupt corrections to both inputs as the local thermal condition changes, whereas the policy modulates them smoothly across each track. In the final CFD temperature fields, the unoptimized scan develops a large high-temperature region associated with excessive penetration and accumulated heat, whereas both controlled cases retain a localized melt region (Fig.~\ref{fig:res_control}c).

\subsection{Generalization and computational efficiency}
\label{sec:res_policy}

\begin{figure}[!tbp]
  \centering
  \includegraphics[width=\textwidth]{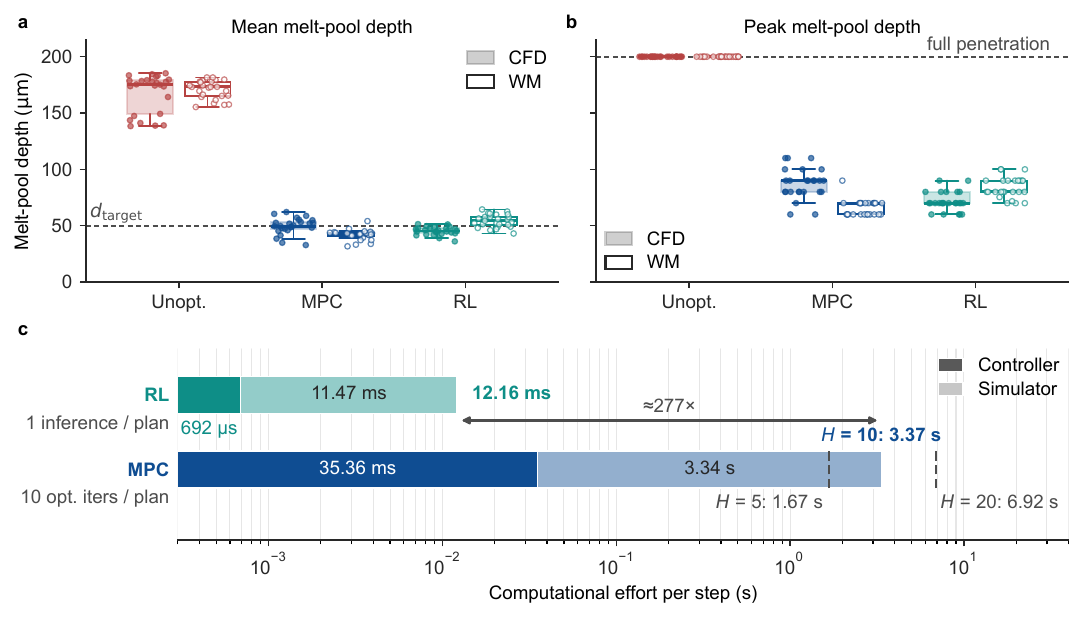}
  \caption{\textbf{Generalization and computational benchmark across unseen overhang cases.} \textbf{(a,b)} Distributions of mean and peak melt-pool depth for the unoptimized scan, MPC and RL across the 30 cases. World-model predictions (open) are compared with CFD re-simulations (filled) to assess generalization under control-induced trajectories. Dashed lines indicate the $50~\mu\mathrm{m}$ target depth in \textbf{(a)} and the full-penetration threshold in \textbf{(b)}. Both controllers bring the mean depth toward the target and suppress the large peak depths observed in the unoptimized scans. The WM under-predicts the depths under MPC and over-predicts those under RL. \textbf{(c)} Computational cost per decision for RL (one control interval) and MPC (one replanning cycle, covering two control intervals), separated into controller and surrogate-rollout time. RL requires $692~\mu\mathrm{s}$ for policy inference and $11.47~\mathrm{ms}$ for the surrogate rollout ($12.16~\mathrm{ms}$ total), whereas MPC with $H=10$ requires $35.36~\mathrm{ms}$ for optimization and $3.34~\mathrm{s}$ for repeated surrogate rollouts ($3.37~\mathrm{s}$ total). All latency measurements were obtained on a single NVIDIA GeForce RTX 5090 GPU.}
  \label{fig:res_benchmark}
\end{figure}

Across $30$ overhang cases with initial conditions drawn from the constant-parameter and transient test sets (Methods), the WM reproduces the shift of mean and peak depth from the unoptimized scans toward the target, but with a controller-dependent bias: it under-predicts the depths reached under MPC and over-predicts those reached under RL (Fig.~\ref{fig:res_benchmark}a,b). MPC often moves power and speed between their bounds within a single control interval, exceeding the per-step variations seen during fine-tuning ($60~\mathrm{W}$ and $0.4~\mathrm{m\,s^{-1}}$) and entering the rapidly varying regime where transition errors are largest (Fig.~\ref{fig:res_stepper}h-l). Because MPC optimizes directly through the model, it can also favour actions for which the model under-predicts depth, a form of model exploitation that biases its predictions downward. In CFD, both controllers nonetheless bring the mean depth close to the $50~\mu\mathrm{m}$ target and substantially reduce the full-penetration peaks of the unoptimized scans. Five further cases, drawn at random from the $30$, are detailed in Figs.~\ref{fig:extended_control_v0_p0}-\ref{fig:extended_control_v2_p4}.

The two approaches differ substantially in cost (Fig.~\ref{fig:res_benchmark}c). The policy requires $692~\mu\mathrm{s}$ for inference and $11.47~\mathrm{ms}$ for the surrogate rollout, or $12.16~\mathrm{ms}$ per $50~\mu\mathrm{s}$ control interval, whereas MPC with $H=10$ requires $3.37~\mathrm{s}$ per replanning cycle of $N=2$ intervals, or about $1.69~\mathrm{s}$ per interval. Amortization therefore makes each control decision approximately $140$ times faster.

\section{Discussion}
\label{sec:discussion}

This study shows that a WM with generative latent dynamics can connect high-fidelity thermal-fluid simulation to control of transient three-dimensional melt pools. The latent plant remains stable under recursive action-conditioned rollouts, supports both gradient-based receding-horizon optimization and policy learning in imagination, and yields schedules that retain their effect when replayed in CFD across unseen overhang geometries.

The ablations separate the roles of representation and transition. Finite scalar quantization provides a compact, bounded code at a reconstruction cost several-fold to an order of magnitude below the rollout error, so the accuracy of the WM is governed by its dynamics rather than by its representation. The deterministic regressors reach lower one-step error but degrade under recursion, whereas the generative transition trades a small one-step error for rollouts that remain accurate over the horizon, most clearly under rapidly varying inputs. We attribute this to conditional-mean predictions leaving the manifold of realizable states, as expected when temperature and phase morphology do not determine the hidden flow, although the present experiments do not isolate this mechanism. Because the noise seed is fixed so that each planning rollout is a single differentiable sample, the generative transition currently benefits control through more accurate rollouts rather than through its predicted spread. Risk-aware objectives averaging over several samples could exploit this spread, which may become important when control objectives extend beyond penetration depth to shape-, defect-, or uncertainty-sensitive criteria.

CFD re-simulation provides the strongest evidence that the WM is suitable for control, because an optimizer can exploit surrogate errors and produce schedules that work only within the learned dynamics. The WM continues to capture the CFD response after control shifts the state and action distributions, although with controller-dependent errors consistent with partial exploitation of model error by MPC. Both controllers nonetheless suppress full penetration in CFD while regulating mean depth near the target. This does not establish transfer to an experimental plant, but it shows that CFD can serve both as the source of training data and as a high-fidelity filter for control schedules before physical deployment.

The control resolution also differs from the layer-level \cite{liao-mcpherson_layer--layer_2025,KAVAS2026105124} and vector-level \cite{kirschbaum2025vector} approaches used to make LPBF control tractable. Both controllers update laser power and scan speed as the heat source advances along an active scan vector, enabling in-line control within an individual track in silico. This resolution matters for overhangs, where support conditions and accumulated heat vary over distances shorter than a scan track \cite{molnar_situ_2021,ashby_thermal_2022}, and regulating depth between a minimum bonding depth and the available support thickness targets both lack of fusion and burn-through.

The framework fits the existing control hierarchy of an LPBF machine, in which a scan controller executes time-ordered position, laser-power, and trigger commands through synchronized galvanometer and laser interfaces, and in-line monitoring adds a measurement path from a coaxial camera or photodetector \cite{craeghs_feedback_2010,wang_real-time_2023,yeung_toward_2025}. A measurement-conditioned encoder would estimate the latent state, and the policy would return power and speed set-points through the existing command interface, leaving motion, synchronization, and safety to the native controllers. Laser power offers the most direct route, whereas within-track speed modulation must also respect galvanometer acceleration and laser-position synchronization limits. Because each transition advances the process by \(\Delta t=50~\mu\mathrm{s}\), the policy is about \(240\) times, and MPC about \(3.4\times10^{4}\) times, slower than real time on a single RTX 5090 GPU, so the results demonstrate millisecond-scale predictive control rather than a real-time feedback loop at the melt-pool timescale. The model timestep need not match the machine control interval, however: action chunking \cite{zhao_learning_2023}, and real-time chunking demonstrated for flow-based policies \cite{black_realtime_2025}, would let the controller run less often while the hardware executes blocks of high-rate commands. Because the transition, rather than action generation, dominates the latency, one-step distillation of the flow-matching transition and compiler-level optimization would reduce the decision time itself.

The demonstration remains entirely in silico. The test cases probe new conditions only within the simulated space and do not establish transfer across materials, machines, or discrepancies between CFD and experiment. The transition is conditioned on absolute heat-source coordinates within the \(1~\mathrm{mm}\times1~\mathrm{mm}\) domain, so part-scale scan paths would require a moving window attached to the heat source. The controller also assumes a latent state derived from three-dimensional temperature and phase fields, whereas a machine provides indirect and often surface-weighted measurements. For this reason, the unoptimized scan serves as the reference here, and a benchmark against reactive feedback laws acting on measurable surface signals is deferred until a sensing pathway links such measurements to the latent state. Joint embeddings of top-view thermal-camera observations with high-fidelity three-dimensional melt-pool simulation offer a concrete route to this encoder \cite{zhang_scientific_2026}, which, together with calibration against the physical process, safeguards outside the training distribution, and testing on an open LPBF platform, is the key remaining step towards machine-integrated in-line control.

Three elements of the approach are not specific to LPBF: pretraining on quasi-steady forcing before adapting to transient forcing, a generative transition over a partially observed state, and re-simulation of every optimized schedule in the solver that generated the training data as a check against exploitation of surrogate errors. They apply to other expensive multiphysics processes controlled through a few actuators, such as welding and directed energy deposition, although each requires process-specific data and validation.
\section{Methods}
\label{sec:methods}

\subsection{Branched FSQ autoencoder}
\label{sec:methods_autoencoder}

The autoencoder (Fig.~\ref{fig:wm_details}, top) converts the partially observed melt-pool state $\mathbf{X}_t=(\mathbf{T}_t,\boldsymbol{\alpha}_t)$ into a compact discrete latent representation. Because similar spatial patterns can occur at different physical magnitudes, we factorize the encoder output into a morphology component shared by both fields and two field-specific amplitude components. Following the structure-amplitude principle introduced for physical-field tokenization \cite{lingsch_phaedra_2026}, the continuous latent representation is written as
\begin{equation}
  \mathbf{z}_t=\mathcal{E}(\mathbf{T}_t,\boldsymbol{\alpha}_t)=
  \left[\mathbf{z}_t^{\mathrm{morph}},\mathbf{z}_t^{T},
  \mathbf{z}_t^{\alpha}\right]
  \label{eq:latent_split}
\end{equation}
where $\mathbf{z}_t^{\mathrm{morph}}$ represents spatial patterns shared by the temperature and phase-indicator fields, whereas $\mathbf{z}_t^{T}$ and $\mathbf{z}_t^{\alpha}$ represent their respective amplitudes.

Finite scalar quantization (FSQ) \cite{mentzer_finite_2023} is applied component-wise using quantization resolutions appropriate to the different latent components,
\begin{equation}
  \mathbf{z}_t^{\mathrm{q}}=\mathcal{Q}_{\mathrm{FSQ}}(\mathbf{z}_t)=
  \left[\mathcal{Q}_{\mathrm{morph}}(\mathbf{z}_t^{\mathrm{morph}}),
  \mathcal{Q}_{T}(\mathbf{z}_t^{T}),
  \mathcal{Q}_{\alpha}(\mathbf{z}_t^{\alpha})\right]=
  \left[\mathbf{z}_t^{\mathrm{q,morph}},\mathbf{z}_t^{\mathrm{q},T},
  \mathbf{z}_t^{\mathrm{q},\alpha}\right]
  \label{eq:tokenizer}
\end{equation}
Restricting the morphology component to a finite set of scalar combinations creates an implicit vocabulary of reusable spatial patterns without a learned codebook. The amplitude components are quantized more finely to retain field-specific magnitude information. This factorization biases related melt-pool states toward sharing discrete morphology patterns while allowing their temperature and phase amplitudes to vary independently. It therefore provides the transition model with a structured and regularized state space. Its reconstruction cost relative to a continuous bottleneck is quantified in Fig.~\ref{fig:res_fsq}a,b.

The decoder branches reconstruct the physical fields from the relevant components of the same aggregate state $\mathbf{z}_t^{\mathrm{q}}$,
\begin{equation}
  \hat{\mathbf{T}}_t=\mathcal{D}_{T}\left(\mathbf{z}_t^{\mathrm{q,morph}},
  \mathbf{z}_t^{\mathrm{q},T}\right),\qquad
  \hat{\boldsymbol{\alpha}}_t=\mathcal{D}_{\alpha}
  \left(\mathbf{z}_t^{\mathrm{q,morph}},\mathbf{z}_t^{\mathrm{q},\alpha}\right),
  \qquad
  \hat{\boldsymbol{\mu}}_t=\mathcal{M}(\hat{\mathbf{T}}_t,
  \hat{\boldsymbol{\alpha}}_t)
  \label{eq:branched_decoder}
\end{equation}
Both branches use the same representation of the coupled spatial structure but retain separate amplitude information and field-specific reconstruction mappings. The melt-pool field $\hat{\boldsymbol{\mu}}_t$, the liquid-metal fraction, is computed from the reconstructed temperature and phase-indicator fields rather than learned as an independent output. The map $\mathcal{M}$ multiplies the phase indicator by a smooth $\tanh$ step of temperature across the melting interval, mirroring the melt-fraction definition of the CFD solver (Supplementary Information, Eq.~S5). The encoder receives neither the laser controls nor the unobserved CFD variables. The controls enter through the latent transition model.

\subsection{Latent flow-matching dynamics}
\label{sec:methods_dynamics}

Given a past window $\mathbf{z}^{\mathrm{q}}_{t-h:t}$, the transition is conditioned on the heat-source position at the start of the interval and the laser power applied during it, $\mathbf{c}_t=(\mathbf{x}^{L}_t,P_t)$. It represents the next state through
\begin{equation}
  \mathbf{z}^{\mathrm{q}}_{t+1}\sim p_{\theta}\!\left(
  \mathbf{z}^{\mathrm{q}}_{t+1}\mid
  \mathbf{z}^{\mathrm{q}}_{t-h:t},\mathbf{c}_t\right).
  \label{eq:transition}
\end{equation}
The transition advances the latent state over a fixed interval $\Delta t=50~\mu\mathrm{s}$ and does not receive scan velocity as a separate scalar input. Instead, the heat-source coordinates satisfy $\mathbf{x}^{L}_{t+1}-\mathbf{x}^{L}_t=v_t\Delta t\,\mathbf{e}_t$, where $\mathbf{e}_t$ is the tangent to the prescribed scan path. The coordinate sequence therefore encodes the heat-source displacement, and hence the scan speed and direction, with a delay of one interval: the speed applied during interval $t$ determines $\mathbf{x}^{L}_{t+1}$ and first enters the model through the condition of the next transition.

The temperature and phase-indicator fields provide only a partial observation of the thermal-fluid system because the underlying velocity and pressure fields are not included. Multiple future states may therefore be compatible with the same observed history and action. A deterministic model trained with a pointwise loss approximates their conditional mean, whereas the generative transition represents the conditional distribution.

The transition is parameterized by a diffusion transformer (DiT) \cite{peebles_scalable_2023} trained with flow matching \cite{lipman_flow_2023} on the latent residual $\mathbf{r}_t=\mathbf{z}^{\mathrm{q}}_{t+1}-\mathbf{z}^{\mathrm{q}}_t$. Flow matching is adopted because its linear probability path supports deterministic few-step sampling with a simple Euler integrator, which keeps each transition inexpensive and differentiable inside the recursive rollouts required for gradient-based control and policy training. Given Gaussian noise $\mathbf{x}_0\sim\mathcal{N}(\mathbf{0},\mathbf{I})$ and flow-matching time $\tau\sim\mathcal{U}(0,1)$, the linear probability path and target velocity are
\begin{equation}
  \mathbf{x}_{\tau}=(1-\tau)\mathbf{x}_0+\tau\mathbf{r}_t,
  \qquad \mathbf{v}_{\tau}=\mathbf{r}_t-\mathbf{x}_0.
  \label{eq:fm_path}
\end{equation}
The noisy residual is concatenated with the past latent window and passed to the DiT, while $\mathbf{c}_t$ and $\tau$ condition its adaptive layer-normalization blocks. The network minimizes
\begin{equation}
  \mathcal{L}_{\mathrm{FM}}=\mathbb{E}\!\left[
  \left\|\mathbf{v}_{\theta}\!\left(\mathbf{x}_{\tau},
  \mathbf{z}^{\mathrm{q}}_{t-h:t},\mathbf{c}_t,\tau\right)
  -\mathbf{v}_{\tau}\right\|_2^2\right]
  \label{eq:fm}
\end{equation}
The concatenated latent input is divided into anisotropic three-dimensional patches before self-attention. The in-plane directions are pooled into attention tokens, whereas the depth direction retains its full latent resolution because melt-pool penetration depth is the principal controlled quantity. The control condition and flow-matching time modulate the attention and multilayer-perceptron blocks through conditional normalization. At rollout time, a four-step Euler sampler integrates the predicted velocity $\mathbf{v}_{\theta}$ into the latent residual, so that each transition requires only four network evaluations. The residual is added to $\mathbf{z}^{\mathrm{q}}_t$ to yield $\mathbf{z}^{\mathrm{q}}_{t+1}$, which can optionally be re-quantized by FSQ (Fig.~\ref{fig:wm_details}, bottom). With re-quantization, gradients pass through the rounding during gradient-based control and policy training with the straight-through estimator used to train FSQ \cite{mentzer_finite_2023}. To reduce optimization cost, the MPC planning rollouts instead use two Euler steps per transition (Supplementary Information, Section~S7).

\subsection{Training strategy and generalization protocol for world model}
\label{sec:methods_training}

The autoencoder and transition model are trained sequentially, and the autoencoder is trained on individual high-fidelity LPBF frames only, never on transitions. It is first pretrained on frames of the constant-schedule cases. This pretrained autoencoder is the FSQ arm of the representation ablation, compared against a continuous bottleneck trained on the same frames (Fig.~\ref{fig:res_fsq}a,b). It is then fine-tuned on frames of time-varying random-walk cases, so that the representation also covers melt-pool states induced by changing inputs. This fine-tuned autoencoder is frozen and used by the WM: it encodes the simulated trajectories into the fixed latent representation on which every transition model, including the deterministic baselines, is trained. The action-conditioned transition model is trained on these latent sequences together with their heat-source controls. Each simulated case is generated with a randomly initialized substrate overhang geometry, residual-temperature field and scan path, so that cases differ in local support thickness, thermal history and trajectory. The detailed information can be found in Supplementary Information, Section~S2. Generalization is evaluated on complete simulation trajectories excluded from training in their entirety, with held-out cases containing previously unseen combinations of residual-temperature fields, hatching paths, substrate overhang geometries and laser-control histories. The rollout-fidelity evaluation uses $25$ such cases with constant parameters (Fig.~\ref{fig:res_surrogate}), the representation comparison uses these and the transient cases below (Fig.~\ref{fig:res_fsq}a,b), whereas the transition ablation uses $88$ test cases with time-varying controls in four regimes, namely $30$ step, $20$ ramp, $19$ modulated and $19$ fast random-walk schedules (Fig.~\ref{fig:res_stepper}c-l). To adapt the dynamics to time-varying inputs, the transition model pretrained on constant schedules is fine-tuned on $n=10$-$100$ time-varying cases, drawn in equal parts from a pool of $400$ random-walk and $400$ banded-schedule cases, with early stopping on a shared $40$-case validation set from the same pool. The model fine-tuned on $n=100$ cases is used by both controllers. A Pure Transient reference with the same architecture and training recipe as the base model is trained from scratch on all $800$ cases of this pool, including the fine-tuning and validation cases, with a random $90/10$ split of its latent windows for early stopping. Complete architecture specifications, dataset compositions and training hyperparameters are provided in the Supplementary Information (Sections~S2-S6).

\subsection{Control formulation and controller implementation}
\label{sec:methods_control}

The learned WM turns melt-pool control into finite-horizon planning over a differentiable latent plant (Fig.~\ref{fig:controllers}). The hatching path is fixed in advance, and the controller determines the laser power and travel speed along that path. At control step $t$, it selects the bounded action $\mathbf{a}_t=(v_t,P_t)$, with $v_t\in[v_{\min},v_{\max}]$ and $P_t\in[P_{\min},P_{\max}]$. The current path coordinate $s_t$ determines the heat-source position $\mathbf{x}^{L}_t=\boldsymbol{\gamma}(s_t)$ and hence the transition condition $\mathbf{c}_t=(\mathbf{x}^{L}_t,P_t)$. The frozen WM then advances the latent state according to $\mathbf{z}^{\mathrm{q}}_{t+1}\sim p_{\theta}(\mathbf{z}^{\mathrm{q}}_{t+1}\mid\mathbf{z}^{\mathrm{q}}_{t-h:t},\mathbf{c}_t)$, while the speed component advances the path coordinate as $s_{t+1}=s_t+v_t\Delta t$. Here, $\boldsymbol{\gamma}$ denotes the prescribed spatial scan path. Consistent with the displacement encoding above, scan velocity is not an additional component of $\mathbf{c}_t$. Instead, it determines the heat-source displacement between successive position conditions over the fixed interval $\Delta t$. The decoder is queried when a physical output such as melt-pool depth is needed. Melt-pool depth is measured from the initial substrate top surface to the deepest point at which the decoded liquid fraction reaches $0.5$. For evaluation, this point is located with sub-voxel interpolation, whereas the control objectives use a smooth relaxation of the threshold depth that is differentiable with respect to the decoded fields.

Both controllers compare the decoded melt-pool depth $d_k$ with a target that is capped by the local support thickness $\delta_k$ below the laser. With a nominal fusion depth $d_{\mathrm{fuse}}$, a minimum bonding depth $d_{\mathrm{bond}}$ and a relative-depth limit $\rho^{*}$, the target at step $k$ is
\begin{equation}
  d^{*}_{k}=\min\!\left(d_{\mathrm{fuse}},\;\min\!\bigl(\rho^{*}\delta_k,\;\max(d_{\mathrm{bond}},\;\rho^{*}\bar{\delta}_k)\bigr)\right),
  \label{eq:target}
\end{equation}
where $\bar{\delta}_k$ is the support thickness a short distance ahead of the laser along the path. Away from the overhang, the target rests at $d_{\mathrm{fuse}}$. Over thin support, it is capped so that the melt pool penetrates at most a fraction $\rho^{*}$ of the available thickness, and the lookahead lowers it, down to $d_{\mathrm{bond}}$, before the laser reaches thinner support. Model-predictive control (MPC) minimizes the finite-horizon objective
\begin{equation}
  J_t^{(H)}=\sum_{k=t}^{t+H-1}\ell_k,\qquad
  \ell_k=w_{\mathrm{band}}\left(\frac{d_k-d^{*}_{k}}{\sigma_{\mathrm{band}}}\right)^{2}+w_{\mathrm{var}}\left(\frac{d_k-d_{k-1}}{d_{\mathrm{fuse}}}\right)^{2},
  \label{eq:objective}
\end{equation}
where the sum runs over the laser-on steps of the window. The first term penalizes departures from the thickness-capped target, and therefore both insufficient fusion and excessive penetration relative to the support. The second regularizes step-to-step changes in depth. The target is built from thickness lookups that are held constant during differentiation, so that the optimizer cannot lower it by changing the speed. Parameter values are listed in the Supplementary Information (Section~S7). At each replanning step, the gradient-based controller optimizes the candidate control sequence $\mathbf{C}_t=(\mathbf{a}_t,\ldots,\mathbf{a}_{t+H-1})$ directly through the frozen surrogate (Fig.~\ref{fig:controllers}a), applies the first $N\le H$ actions of the optimized sequence $\mathbf{C}_t^{*}$, and then resolves the length-$H$ problem from the updated latent state.

The amortized SHAC controller uses the same physical objective as MPC, defining the reward $r_{b,t+j}=-\ell_{t+j}$ as the negative stage cost of Eq.~\eqref{eq:objective} at step $t+j$ of imagined trajectory $b$. It learns an actor $\pi_\psi$ and value critic $V_\phi$ from trajectories generated by the same frozen WM. The observation $\mathbf{o}_t$ contains the quantized latent history together with the laser position and path direction, previous action, scan progress and local-support context. For each of $B$ sampled initial observations, the actor maps $\mathbf{o}_{b,t+j}$ to a bounded action $\mathbf{a}_{b,t+j}=\pi_\psi(\mathbf{o}_{b,t+j})$, and the WM advances the resulting trajectory for $L$ steps. The bootstrapped return and actor loss are
\begin{equation}
  G_b^{(L)}=
  \sum_{j=0}^{L-1}\beta^j r_{b,t+j}
  +\beta^L V_\phi(\mathbf{o}_{b,t+L}),
  \qquad
  \mathcal{L}_{\mathrm{actor}}(\psi)
  =-\frac{1}{B}\sum_{b=1}^{B}G_b^{(L)}
  \label{eq:shac_actor}
\end{equation}
where $\beta$ is the discount factor. During the actor update, $\phi$ is held fixed, while gradients from both the stage rewards and terminal value are propagated through the actor and the complete differentiable rollout.

The critic targets are generated backwards along each imagined trajectory,
\begin{equation}
  R_{b,t+L}=\operatorname{sg}\!\left[V_\phi(\mathbf{o}_{b,t+L})\right],
  \qquad
  R_{b,t+j}=r_{b,t+j}+\beta R_{b,t+j+1}
  \label{eq:shac_targets}
\end{equation}
and the critic is fitted by
\begin{equation}
  \mathcal{L}_{\mathrm{critic}}(\phi)
  =\frac{1}{BL}\sum_{b=1}^{B}\sum_{j=0}^{L-1}
  \left[V_\phi(\mathbf{o}_{b,t+j})-\operatorname{sg}\!\left[R_{b,t+j}\right]\right]^2
  \label{eq:shac_critic}
\end{equation}
where $\operatorname{sg}$ denotes stop-gradient. The observations and return targets used for this critic fit are detached from the world-model computation graph.

This implementation retains the defining SHAC principle of combining truncated differentiable rollouts with a terminal critic \cite{xu2021accelerated}, but differs from the original formulation in three respects: the actor is deterministic and bounded, the critic uses the full truncated return rather than a TD-$\lambda$ mixture of multiple $k$-step returns, and terminal bootstrapping uses the current critic rather than a slowly updated target critic. These simplifications suit the present low-dimensional action space, in which the two bounded controls are scan speed and laser power, while sampled latent initial states and seeded transition samples provide trajectory diversity. The short differentiable world-model rollout permits direct pathwise actor gradients. Using the full return with alternating actor and critic updates avoids additional TD-$\lambda$ and target-network machinery while preserving an estimate of performance beyond the truncation point.

After training, only the actor is retained from the actor-critic pair. At deployment, $\mathbf{a}_t=\pi_{\psi^{*}}(\mathbf{o}_t)$ is evaluated once per control step, the frozen WM advances the latent state, and neither the critic nor a backward optimization pass is required.

\subsection{Training and testing protocol for the control cases}
\label{sec:methods_control_protocol}

The actor and critic are trained entirely within the frozen WM. Each training episode is initialized by sampling a random case and a random time window from the constant-schedule training trajectories of the WM, so the substrate overhang geometry, temperature initialization and scan path of every episode are drawn at random from its training data. The imagined rollout then evolves from this sampled latent state under the actions of the current actor.

The control benchmark comprises $30$ cases whose initial conditions are drawn at random from the union of the $25$-case constant-parameter test set and the $88$-case transient test set, neither of which is used for training (Fig.~\ref{fig:res_benchmark}). Each case is executed with the unoptimized schedule, MPC and the amortized policy, and every controlled trajectory is re-simulated in CFD. The controller hyperparameters and the re-validation protocol are provided in the Supplementary Information (Section~S7).

\subsection{CFD simulation}
\label{sec:methods_cfd}

All training and evaluation data are generated with a high-fidelity thermal-fluid model of LPBF, implemented as a customized two-phase solver in OpenFOAM~v5 \cite{weller_tensorial_1998} that extends the volume-of-fluid (VOF) solver interFoam \cite{hirt_volume_1981}. The same process model was employed in the previous work~\cite{muther_identifying_2025}, where its predictions were validated against operando synchrotron tomographic microscopy. Here, the solver is parameterized for Ti-6Al-4V. The metal and the surrounding argon atmosphere are treated as incompressible Newtonian phases, with the gas-metal interface advected by the geometric isoAdvector scheme \cite{roenby_computational_2016}. The metal volume fraction of this formulation is the phase-indicator field $\boldsymbol{\alpha}$ observed by the WM. The solver resolves the dominant melt-pool physics, including Marangoni shear, evaporation-induced recoil pressure, buoyancy, mushy-zone damping and interfacial heat losses, under a moving Gaussian laser source with absorptivity $\eta=0.5$ and beam radius $R_l=50~\mu\mathrm{m}$. The source terms, the Ti-6Al-4V thermophysical properties and the remaining model constants are given in the Supplementary Information (Section~S1 and Supplementary Tables~S1 and~S2).

The computational domain of $1~\mathrm{mm}\times1~\mathrm{mm}\times0.32~\mathrm{mm}$ is discretized by a uniform Cartesian grid of $100\times100\times32$ cells, giving the $10~\mu\mathrm{m}$ cell size at which the WM observes the fields. The substrate occupies the lowest $0.2~\mathrm{mm}$ and argon fills the region above. All domain boundaries are stationary no-slip, adiabatic walls, so that heat leaves the metal only through the interfacial convective, radiative and evaporative sinks. Pressure-velocity coupling uses the PIMPLE algorithm, and the time step (initially $0.5~\mu\mathrm{s}$) adapts to keep the Courant number below $0.75$ and the interface Courant number below $1$.

Each case is initialized with a randomized substrate overhang geometry, surface morphology and residual-heat field, and is scanned along a hatching path of random orientation with track spacings of $50$-$200~\mu\mathrm{m}$, with the laser switched off for $0.25~\mathrm{ms}$ while repositioning between tracks. The full randomization protocol is given in the Supplementary Information (Section~S2). Training cases apply either constant parameters drawn from a $20\times20$ grid spanning scan speeds of $0.5$-$2.4~\mathrm{m\,s^{-1}}$ and laser powers of $40$-$325~\mathrm{W}$, or time-varying schedules in which both parameters follow a bounded random walk over the same ranges, resampled every $50~\mu\mathrm{s}$ with per-step changes limited to $60~\mathrm{W}$ and $0.4~\mathrm{m\,s^{-1}}$, either across the full ranges or confined to a band of $\pm50~\mathrm{W}$ and $\pm0.5~\mathrm{m\,s^{-1}}$ around the nominal values. The test cases use parameter combinations off the training grid and previously unseen control histories, and a separate transient test set applies step, ramp, modulated and fast random-walk schedules, the last being the random walk above with doubled per-step limits (Supplementary Information, Section~S2). The controlled trajectories are re-simulated by imposing the selected power and speed schedules as piecewise-constant inputs over successive $50~\mu\mathrm{s}$ intervals. Each training and test case is simulated for $4~\mathrm{ms}$ of process time, whereas each control case runs until its predefined scan path has been traversed under the applied schedule, so that its duration depends on the case and the controller. In all cases, cell-centered temperature, velocity, phase-indicator and melt-fraction fields are recorded every $50~\mu\mathrm{s}$, which defines the transition interval $\Delta t$ of the WM. Only the temperature and phase-indicator fields are exposed to the autoencoder. All CFD simulations are performed on the Euler cluster of ETH Zurich, with each case decomposed over $64$ MPI processes on a single AMD EPYC 7763 node. Re-simulating one controlled case required a median wall-clock time of approximately $17$ minutes (interquartile range $13$ to $19$ minutes) on this node, or roughly $10~\mathrm{s}$ of computation per $50~\mu\mathrm{s}$ control interval, compared with $11.47~\mathrm{ms}$ per interval for the world-model rollout on a single GPU (Fig.~\ref{fig:res_benchmark}c). Repeated evaluation of candidate action sequences against the CFD model is therefore infeasible within a control loop, and the WM supplies the fast plant required by both controllers.

\subsection{Statistics}
\label{sec:methods_stats}

Generative models are evaluated with eight rollout samples per case and deterministic models with one, for which the CRPS reduces to the mean absolute error. Melt-pool depth and volume are scored only on laser-on steps. Reported CRPS levels pool all laser-on steps of a case set, so that each step carries equal weight, and their $95\%$ confidence intervals come from a bootstrap that resamples whole cases ($20{,}000$ resamples). Differences between models are computed per case, from each case's mean over its laser-on steps, and averaged over cases, with two-sided $P$ values from a paired bootstrap over the identical case set. Pooled levels and case-averaged differences can therefore differ slightly (Supplementary Information, Section~S6.2).
\section*{Declaration of Competing Interest}
The authors confirm that there are no existing financial conflicts or personal associations that might be perceived as influencing the findings presented in this paper.

\section*{Declaration of Generative AI and AI-assisted technologies in the writing process}
During the preparation of this work, the authors used Claude to improve the readability and language of this article. 
After using this tool, the authors reviewed and edited the content as needed and take full responsibility for the content of the publication.

\section*{Data Availability}
The data supporting the findings of this study have been deposited in the ETH Zurich Research Collection. During peer review, the dataset is available to editors and reviewers through a password-protected repository link \textcolor{blue}{\url{https://libdrive.ethz.ch/index.php/s/WibqAP2gBTfDdW3}}. The data will be made publicly available through the ETH Zurich Research Collection upon publication.

\section*{Code Availability}
The data and code for the project could be found at \textcolor{blue}{\url{https://github.com/yyyyorky/meltpool_world_model}}.
\appendix

\clearpage
\section{Extended Data}
\label{app:extended_data}
\setcounter{figure}{0}

\begin{figure}[H]
  \centering
  \includegraphics[width=0.92\textwidth]{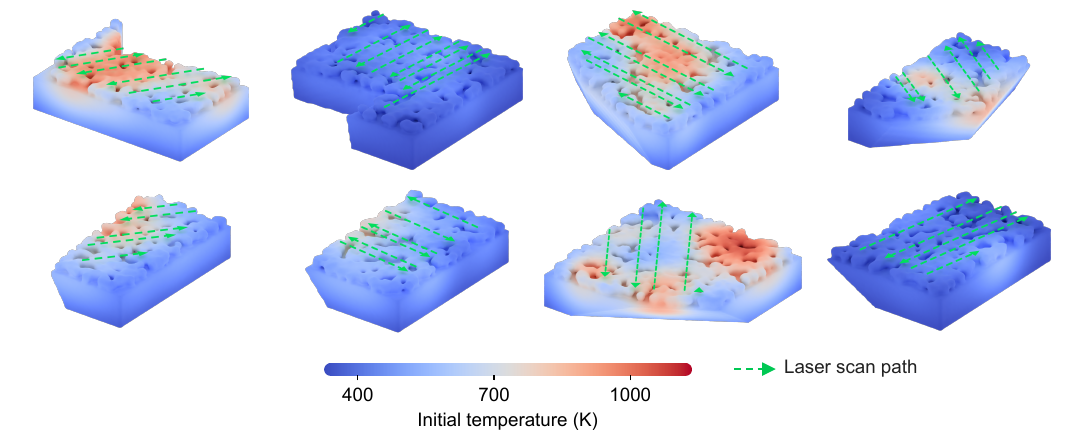}
  \caption{\textbf{Initial conditions and prescribed laser scan paths.} Representative initial temperature fields span distinct residual-heat distributions and substrate geometries. Green dashed arrows indicate the prescribed laser scan paths used to evaluate the world model across these conditions.}
  \label{fig:extended_initial_conditions}

  \vspace{10pt}

  \centering
  \includegraphics[width=0.86\textwidth]{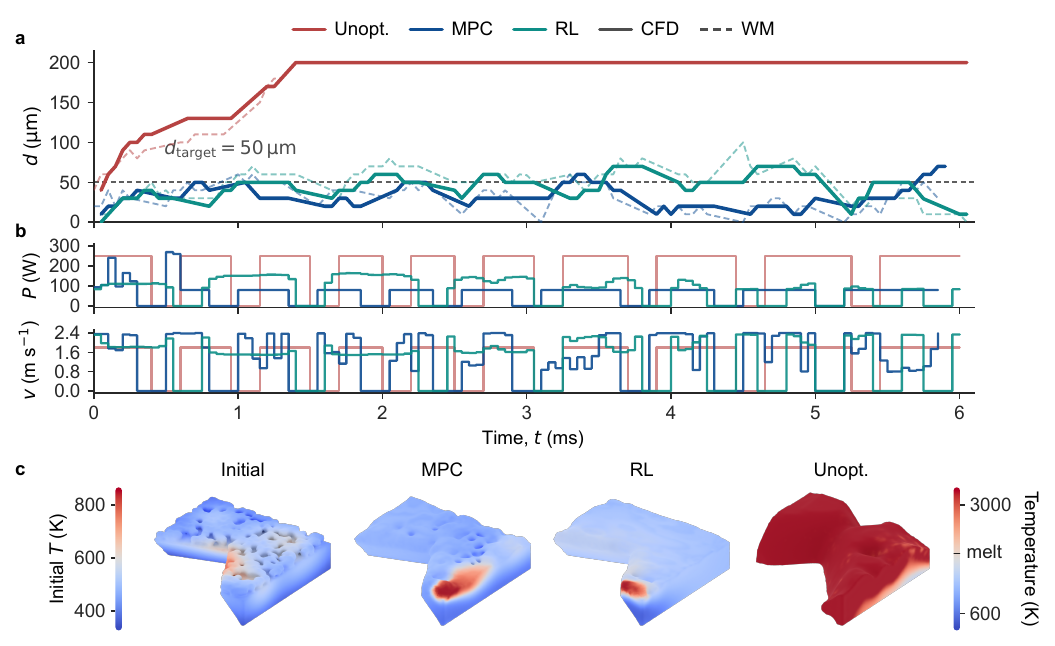}
  \caption{\textbf{Additional world-model control example 1.} \textbf{(a)} Melt-pool depth for the unoptimized scan, MPC and RL, comparing the world-model rollout (dashed) with the corresponding CFD re-simulation (solid). The horizontal dashed line marks the target depth. \textbf{(b)} Laser-power and scan-speed schedules for the three control strategies. \textbf{(c)} Initial temperature field and final CFD temperature fields obtained with MPC, RL and the unoptimized scan.}
  \label{fig:extended_control_v0_p0}
\end{figure}

\clearpage

\begin{figure}[!t]
  \centering
  \includegraphics[width=0.81\textwidth]{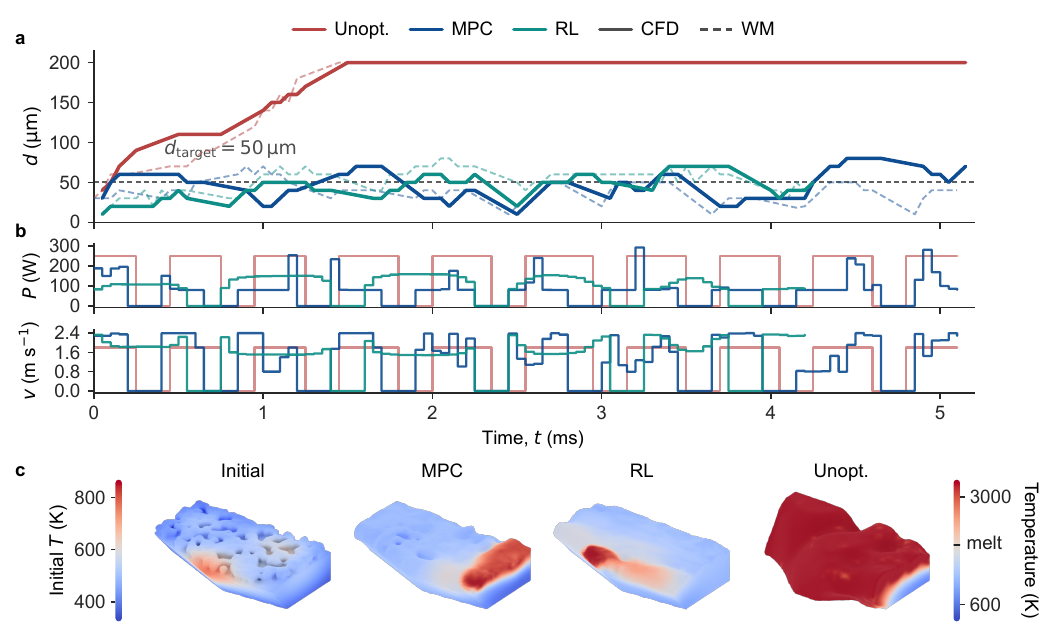}
  \caption{\textbf{Additional world-model control example 2.} \textbf{(a)} Melt-pool depth for the unoptimized scan, MPC and RL, comparing the world-model rollout (dashed) with the corresponding CFD re-simulation (solid). The horizontal dashed line marks the target depth. \textbf{(b)} Laser-power and scan-speed schedules for the three control strategies. \textbf{(c)} Initial temperature field and final CFD temperature fields obtained with MPC, RL and the unoptimized scan.}
  \label{fig:extended_control_v0_p2}

  \vspace{10pt}

  \centering
  \includegraphics[width=0.81\textwidth]{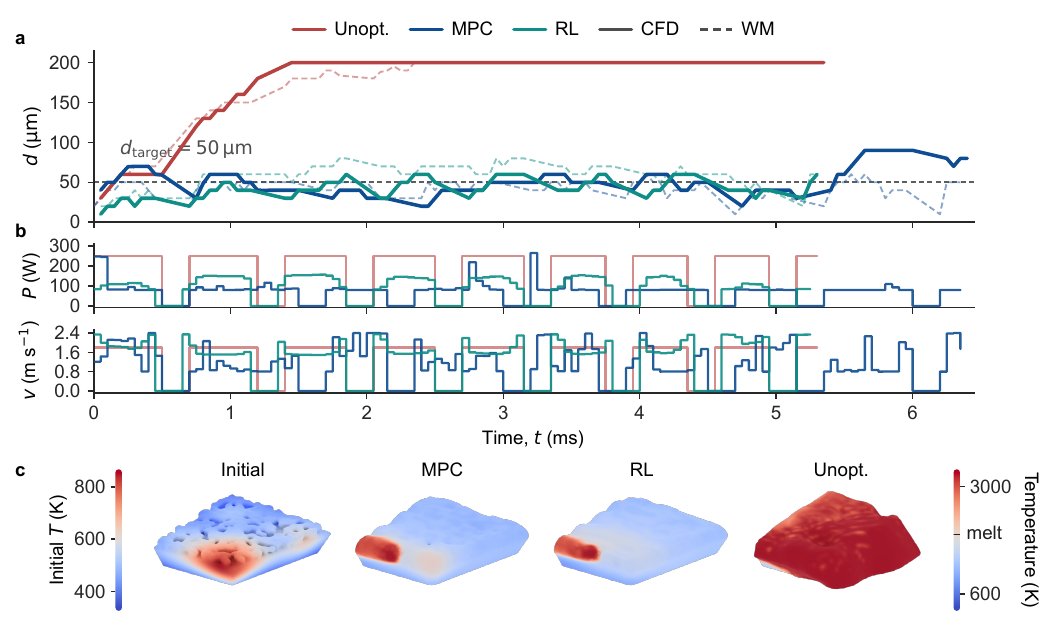}
  \caption{\textbf{Additional world-model control example 3.} \textbf{(a)} Melt-pool depth for the unoptimized scan, MPC and RL, comparing the world-model rollout (dashed) with the corresponding CFD re-simulation (solid). The horizontal dashed line marks the target depth. \textbf{(b)} Laser-power and scan-speed schedules for the three control strategies. \textbf{(c)} Initial temperature field and final CFD temperature fields obtained with MPC, RL and the unoptimized scan.}
  \label{fig:extended_control_v0_p4}
\end{figure}

\clearpage

\begin{figure}[!t]
  \centering
  \includegraphics[width=0.81\textwidth]{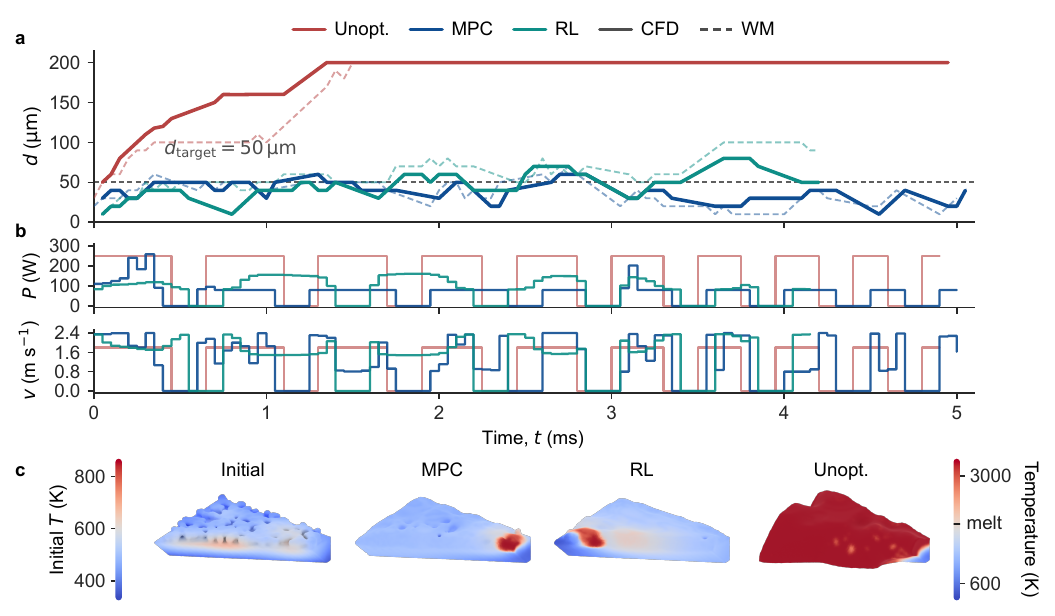}
  \caption{\textbf{Additional world-model control example 4.} \textbf{(a)} Melt-pool depth for the unoptimized scan, MPC and RL, comparing the world-model rollout (dashed) with the corresponding CFD re-simulation (solid). The horizontal dashed line marks the target depth. \textbf{(b)} Laser-power and scan-speed schedules for the three control strategies. \textbf{(c)} Initial temperature field and final CFD temperature fields obtained with MPC, RL and the unoptimized scan.}
  \label{fig:extended_control_v1_p1}

  \vspace{10pt}

  \centering
  \includegraphics[width=0.81\textwidth]{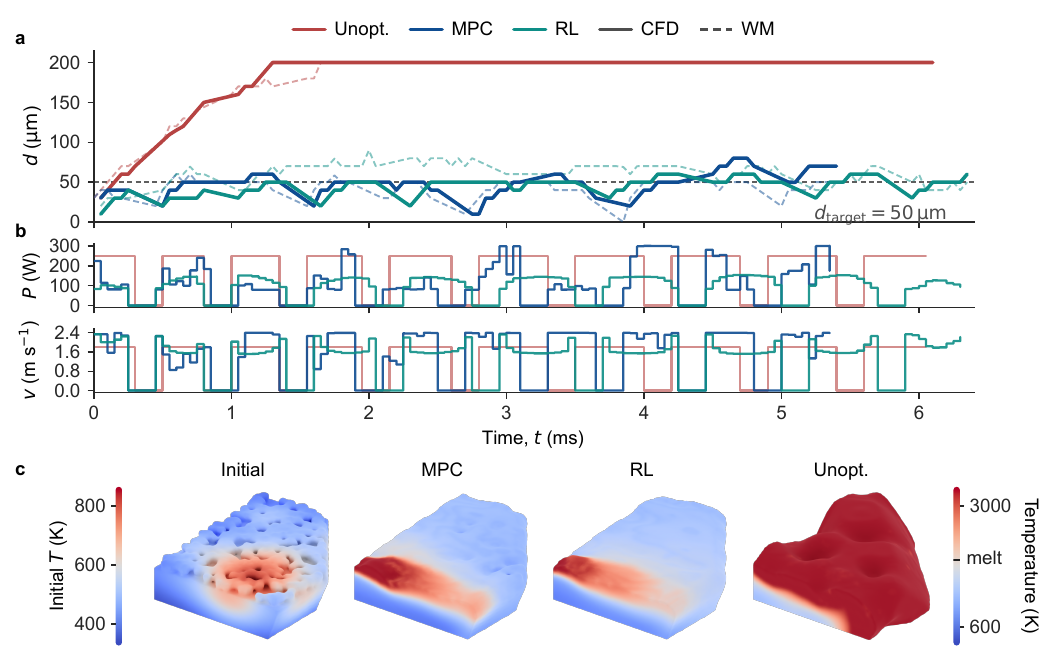}
  \caption{\textbf{Additional world-model control example 5.} \textbf{(a)} Melt-pool depth for the unoptimized scan, MPC and RL, comparing the world-model rollout (dashed) with the corresponding CFD re-simulation (solid). The horizontal dashed line marks the target depth. \textbf{(b)} Laser-power and scan-speed schedules for the three control strategies. \textbf{(c)} Initial temperature field and final CFD temperature fields obtained with MPC, RL and the unoptimized scan.}
  \label{fig:extended_control_v2_p4}
\end{figure}

\clearpage

\clearpage
\bibliography{Bibliothek}

\clearpage


\section*{Supplementary material}

# Supplementary Information for "Generative World Models Enable Predictive Control of Laser Melt Pool Dynamics"

Yiyang Yan[a], Markus Bambach[a], Mohamadreza Afrasiabi[a,*]

[a] *Advanced Manufacturing Lab, ETH Zürich, Leonhardstrasse 21, Zürich 8092, Switzerland*

** Corresponding author: afrasiabi@ethz.ch (M. Afrasiabi)*

This Supplementary Information provides the details required to reproduce the results of the main text: the high-fidelity thermal-fluid model and its numerical solution (Section S1), the randomized case-generation protocol and the composition of all datasets (Section S2), the preparation of the training data (Section S3), the architectures and training protocols of the autoencoder and the transition model (Sections S4 and S5), the baseline models and the evaluation protocol (Section S6), the configuration of the two controllers (Section S7), and the compute environment (Section S8).

## S1 High-fidelity thermal-fluid model

### S1.1 Governing equations and source terms

All training and evaluation data are generated with a customized two-phase solver implemented in OpenFOAM v5 [1], which extends the volume-of-fluid (VOF) framework [2] with the interfacial physics of laser melting. The same process model was validated against operando synchrotron tomographic microscopy in previous work [3]. Here, it is parameterized for Ti-6Al-4V under an argon atmosphere. Both phases are treated as incompressible Newtonian fluids in the laminar regime, and the metal volume fraction of the VOF formulation is the phase-indicator field $\boldsymbol{\alpha}$ observed by the world model. The solver advances the velocity $\mathbf{u}$, pressure $p$, temperature $T$ and phase indicator $\alpha$ according to

$$\nabla \cdot \mathbf{u} = 0, \tag{S1}$$

$$\frac{\partial \alpha}{\partial t} + \nabla \cdot (\alpha \mathbf{u}) = 0, \tag{S2}$$

$$\frac{\partial (\rho \mathbf{u})}{\partial t} + \nabla \cdot (\rho \mathbf{u}\mathbf{u}) = -\nabla p + \nabla \cdot \left[\mu \left(\nabla \mathbf{u} + \nabla \mathbf{u}^{\mathsf{T}}\right)\right] + \mathbf{f}_\sigma + \mathbf{f}_{\mathrm{M}} + \mathbf{f}_{\mathrm{r}} + \mathbf{f}_{\mathrm{b}} + \mathbf{f}_{\mathrm{D}}, \tag{S3}$$

$$\frac{\partial (\rho h)}{\partial t} + \nabla \cdot (\rho \mathbf{u} h) = \nabla \cdot (k \nabla T) + q_{\mathrm{L}} - q_{\mathrm{loss}}, \tag{S4}$$

where the density $\rho$, viscosity $\mu$ and thermal conductivity $k$ are blended between the metal and gas phases by $\alpha$, and the specific enthalpy $h$ includes the latent heats of fusion and evaporation. The momentum sources are surface tension $\mathbf{f}_\sigma$, Marangoni shear $\mathbf{f}_{\mathrm{M}}$, recoil pressure $\mathbf{f}_{\mathrm{r}}$, Boussinesq buoyancy $\mathbf{f}_{\mathrm{b}}$ and the mushy-zone Darcy sink $\mathbf{f}_{\mathrm{D}}$. The energy source $q_{\mathrm{L}}$ is the absorbed laser power

density, and $q_{\text{loss}}$ is the interfacial heat loss. Their closures are summarized below, and the full formulation is given in [3].

The energy equation is solved in enthalpy form with temperature-dependent specific heat and thermal conductivity, blended between solid- and liquid-phase values across the melting interval. The latent heats of fusion and evaporation enter through liquid and vapor fractions that vary smoothly over the solidus-liquidus and boiling ranges, and the local temperature is capped at the boiling temperature. The laser follows the Gaussian irradiance profile $q(r) = \frac{2\eta P}{\pi R_l^2} \exp(-2r^2/R_l^2)$ with absorptivity $\eta = 0.5$ and beam radius $R_l = 50\ \mu$m, and is distributed over the diffuse gas-metal interface in a continuum-surface-force manner [4]. The momentum equation includes surface tension with a linear temperature dependence, the associated Marangoni shear, evaporation-induced recoil pressure [5], Boussinesq buoyancy, and a Kozeny-Carman Darcy sink that suppresses flow in the mushy and solid regions [6]. Heat is extracted from the free surface by convective, radiative and evaporative losses. The thermophysical properties of Ti-6Al-4V are taken from standard references [7, 8] and are listed in Table S1, and the remaining model constants in Table S2.

Table S1: Thermophysical properties of Ti-6Al-4V used in the CFD model [7, 8]. $T$ denotes temperature in K.

| Property | Value |
|---|---|
| Solidus temperature | 1877 K |
| Liquidus temperature | 1923 K |
| Boiling temperature | 3315 K |
| Latent heat of fusion | $2.86 \times 10^5$ J kg$^{-1}$ |
| Latent heat of evaporation | $8.88 \times 10^6$ J kg$^{-1}$ |
| Density | 4420 kg m$^{-3}$ |
| Liquid viscosity | $5 \times 10^{-3}$ Pa s |
| Specific heat, solid | $546 + 0.17T$ J kg$^{-1}$ K$^{-1}$ |
| Specific heat, liquid | 831 J kg$^{-1}$ K$^{-1}$ |
| Thermal conductivity, solid | $6.7 + 0.0114T$ W m$^{-1}$ K$^{-1}$ |
| Thermal conductivity, liquid | 33.4 W m$^{-1}$ K$^{-1}$ |
| Thermal expansion coefficient | $8.6 \times 10^{-6}$ K$^{-1}$ |
| Surface tension at liquidus | 1.65 N m$^{-1}$ |
| Surface-tension temperature coefficient | $-2.6 \times 10^{-4}$ N m$^{-1}$ K$^{-1}$ |

## S1.2 Numerical solution

The computational domain of 1 mm $\times$ 1 mm $\times$ 0.32 mm is discretized by a uniform Cartesian grid of $100 \times 100 \times 32$ cells (10 $\mu$m cell size). The substrate occupies the lowest 0.2 mm and argon fills the region above. All six domain boundaries are stationary no-slip, adiabatic walls, so that heat leaves the metal only through the interfacial convective, radiative and evaporative sinks. Pressure-velocity coupling uses the PIMPLE algorithm with two outer and three pressure-corrector iterations, first-order implicit (Euler) time integration and van Leer-limited divergence schemes for the advected fields. The gas-metal interface is advected with the geometric isoAdvector scheme [9]. The time step starts at 0.5 $\mu$s and adapts to keep the Courant number below 0.75 and the interface Courant number below 1. Each case is decomposed over 64 MPI processes with a hierarchical in-plane

Table S2: Laser, environment and numerical model constants of the CFD model.

| Parameter | Value |
|---|---|
| Absorptivity (laser distribution factor) $\eta$ | 0.5 |
| Beam radius $R_l$ | 50 $\mu$m |
| Ambient temperature | 300 K |
| Convective heat-transfer coefficient | 10 $\mathrm{W\,m^{-2}\,K^{-1}}$ |
| Surface emissivity | 0.4 |
| Argon density | 1.784 $\mathrm{kg\,m^{-3}}$ |
| Argon kinematic viscosity | $1.23 \times 10^{-5}$ $\mathrm{m^2\,s^{-1}}$ |
| Darcy (mushy-zone) coefficient | $10^6$ $\mathrm{kg\,m^{-3}\,s^{-1}}$ |
| Gravitational acceleration | $(0, 0, -9.81)$ $\mathrm{m\,s^{-2}}$ |

decomposition.

# S2 Randomized case generation and datasets

## S2.1 Randomization protocol

Every simulated case is generated from a per-case random seed that determines the substrate geometry, the surface morphology, the initial temperature field, the scan path and, where applicable, the laser schedule. The individual components are constructed as follows.

**Substrate and overhang geometry.** The substrate is a 0.2 mm-thick prism whose top outline is cut by a random polyline of one or two segments and whose bottom outline is offset inward, producing an unsupported overhang whose face is inclined to the vertical. For the pretraining and schedule datasets, the inclination is drawn from $\mathcal{U}(10°, 60°)$, and with probability 0.05 each the cut or the overhang is omitted, yielding occasional fully supported substrates. For the banded-schedule dataset, the inclination is drawn from $\mathcal{U}(45°, 60°)$ and every case contains an overhang. Spherical powder-like features are scattered across the top face: 800 spheres per case with radii drawn from $\mathcal{U}(25, 30)$ $\mu$m and uniformly distributed centers.

**Initial temperature field.** The initial temperature field superposes 30 randomly placed anisotropic Gaussians centered at the substrate surface. Each Gaussian's normalized peak is drawn from $\mathcal{U}(0, 0.25)$, and the summed field $f$ is mapped to $T = 300 + 3700f$ K, giving peak initial temperatures up to approximately 1225 K. Gas cells start at the 300 K ambient temperature.

**Scan path.** Each case is scanned along a hatching pattern of random orientation drawn from $\mathcal{U}(0, \pi)$, with track spacing drawn from $\mathcal{U}(50, 200)$ $\mu$m and an inward margin from the substrate outline drawn from $\mathcal{U}(50, 100)$ $\mu$m. Tracks shorter than 20% of the substrate diagonal are discarded and additional patterns are appended until the prescribed path length is reached. The laser is switched off for 0.25 ms while repositioning between tracks.

**Laser schedules.** Three schedule types are used. (i) *Constant*: laser power and scan speed are held at the case's nominal values, drawn from a $20 \times 20$ grid spanning 0.5-2.4 $\mathrm{m\,s^{-1}}$ (incre-

ments of 0.1 $\mathrm{m\,s^{-1}}$) and 40-325 W (increments of 15 W). (ii) *Random walk*: both parameters are resampled every 50 $\mu$s by a clipped random walk started at the case's nominal grid values, with per-step changes drawn uniformly within $\pm 0.4$ $\mathrm{m\,s^{-1}}$ and $\pm 60$ W and clipped to the global ranges. (iii) *Banded random walk*: as (ii), but each case's walk is additionally confined to a band of $\pm 0.5$ $\mathrm{m\,s^{-1}}$ and $\pm 50$ W around its nominal values, so that the nominal grid point controls the case's processing regime.

**Transient test schedules.** A separate test set probes the response of the world model to time-varying laser inputs that drive the melt pool out of quasi-steady state. Its 90 cases are generated with the substrate, powder, initial-temperature and scan-path protocols above, with nominal power and speed drawn from the $20 \times 20$ training grid and with every power and speed value kept within the global training ranges. The initial set comprises 30 step, 20 ramp, 20 modulated and 20 fast random-walk cases (Fig. S3). (i) *Step*: a single step change is applied at the first laser-on interval at or after $t = 1$ ms, in laser power only (10 cases, $|\Delta P| =$ 30-200 W), in scan speed only (10 cases, $|\Delta v| =$ 0.3-1.5 $\mathrm{m\,s^{-1}}$), or in both (10 cases). In the ten cases with the largest changes, the step coincides with the re-ignition of the laser after a repositioning interval. (ii) *Ramp*: power, speed or both change linearly from their nominal values, starting at $t = 0.5$ ms and lasting 0.5, 1 or 2 ms, with total changes of 40-200 W and 0.4-1.5 $\mathrm{m\,s^{-1}}$ (8 power-only, 6 speed-only and 6 combined cases). (iii) *Modulated*: the power oscillates about its nominal value as a sine or square wave at 0.5, 1, 2 or 5 kHz with an amplitude of 50-100 W, and in 4 of the 20 cases the speed is modulated in phase with an amplitude of 0.3-0.5 $\mathrm{m\,s^{-1}}$. At 5 kHz, one period spans four control intervals. (iv) *Fast random walk*: the clipped random walk of schedule type (ii) with the per-step limits doubled to $\pm 120$ W and $\pm 0.8$ $\mathrm{m\,s^{-1}}$. All seeds are disjoint from those of the training datasets. Two cases, one modulated and one fast random walk, were excluded because their CFD simulations diverged, leaving 88 test cases (30 step, 20 ramp, 19 modulated and 19 fast random walk).

## S2.2 Dataset composition

Table S3 summarizes all datasets. Each case is simulated for 4 ms of process time, except the control-benchmark cases, which run until their predefined scan path has been traversed under the applied schedule. Cell-centered fields are recorded every 50 $\mu$s, which defines the transition interval $\Delta t$ of the world model, so that a 4 ms case contributes 81 frames. The world model is trained in stages. The autoencoder is pretrained on the constant-schedule pretraining dataset, fine-tuned on frames of the random-walk schedule dataset and then frozen. Every transition model is trained on its latents. Representation and dynamics are learned separately: the autoencoder is trained on individual frames and never on transitions, so its fine-tuning exposes only the per-frame representation to time-varying schedules, whereas the base transition model is trained exclusively on constant-schedule trajectories. The comparison between the base and fine-tuned transition models therefore isolates the effect of time-varying data on the learned dynamics. The transition model is first trained on the constant-schedule pretraining dataset, giving the base model of the transition ablation, and is then fine-tuned on 100 time-varying cases. This fine-tuned model is the one used by both controllers. The time-varying cases come from a common pool of 400 random-walk (schedule) and 400 banded-schedule cases, generated with the protocols of Section S2.1. From this pool, 50 cases of each type form the fine-tuning set, with nested subsets of $n =$ 10-75 cases for the data-scaling study, and a further 40 cases form the shared validation set used for early stopping. The Pure Transient reference is trained from scratch on all 800 cases of the pool, including the

fine-tuning and validation cases, and is validated on a random 10% of their latent windows. The 25-case rollout test set uses constant parameters off the training grid: five speeds (0.825, 1.125, 1.425, 1.725 and 1.975 $\mathrm{m\,s^{-1}}$), each lying between two training speeds, crossed with five powers (85, 145, 205, 265 and 315 W), and the 30-case control benchmark draws its initial conditions at random from the union of this rollout test set and the 88-case transient test set, so that the controller-selected schedules are re-simulated from overhang geometries and initial temperature fields unseen during training. All held-out cases are excluded from training in their entirety. The parameter-space coverage of the training and test grids, together with example schedule trajectories, is shown in Fig. S1, and example schedules of the three types in Fig. S2. The 88-case transient test set (Section S2.1) is used for the transition ablation of the main text, and representative schedules of its four regimes are shown in Fig. S3.

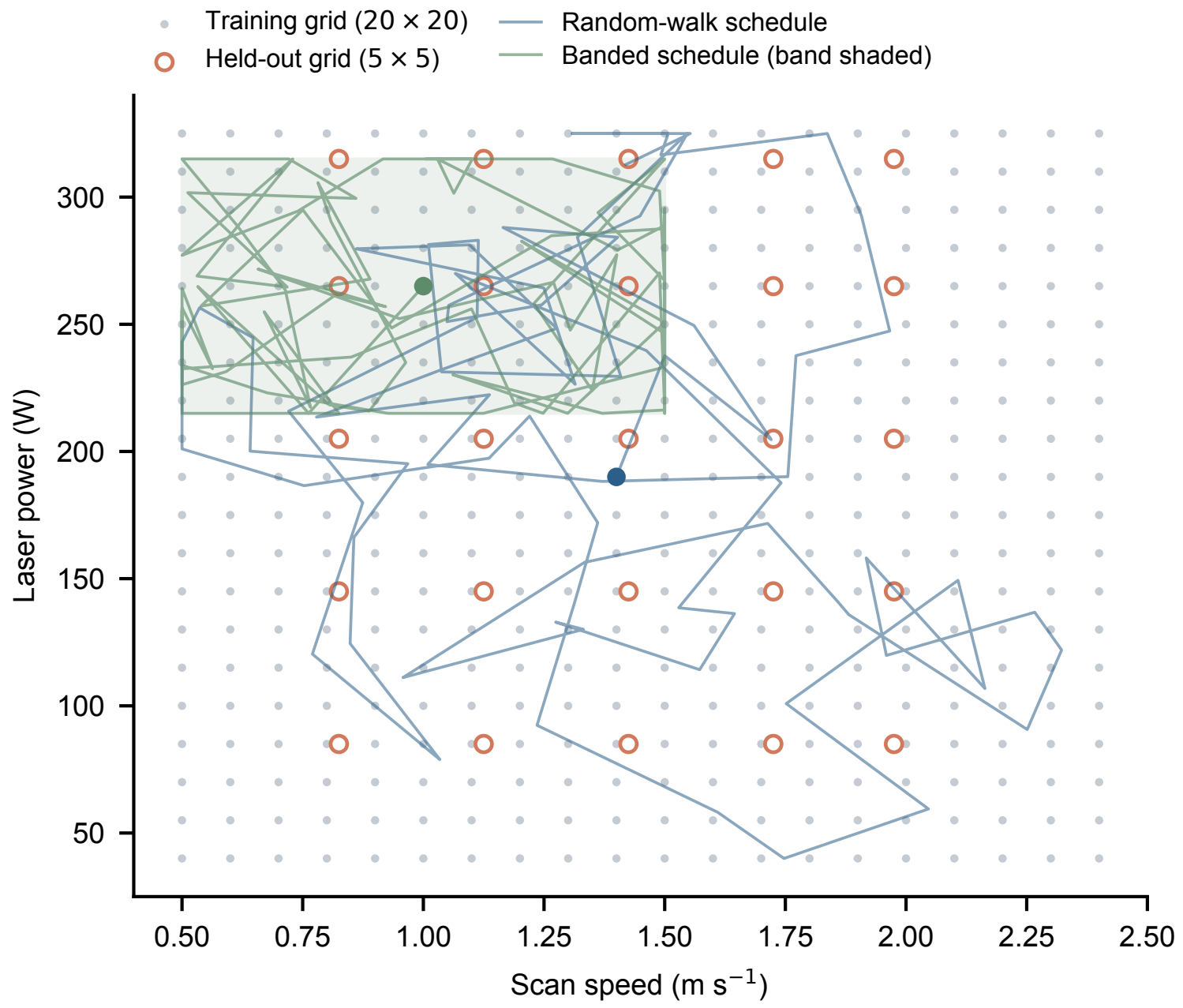


Fig. S1: **Process-parameter coverage of the datasets.** Gray dots mark the $20 \times 20$ constant-parameter training grid and orange circles the 25 test grid points, whose speeds lie between the training values. The blue trace shows an example random-walk schedule trajectory and the green trace an example banded random-walk trajectory, confined to the shaded band of $\pm 0.5$ $\mathrm{m\,s^{-1}}$ and $\pm 50$ W around its nominal point. Filled dots mark the nominal starting points. Trajectories are generated with the sampling rules of Section S2.1.

# S3 Training data preparation

For each recorded frame, the cell-centered temperature, phase-indicator, velocity, and liquid-metal-fraction fields are stored in single-precision HDF5 archives on the $100 \times 100 \times 32$ simulation grid, together with the heat-source position and laser power of every 50 $\mu$s interval. Only the temperature and phase-indicator fields are exposed to the world model. Temperature is normalized as $(T - 300)/3700$ and clipped to $[0, 1]$, corresponding to the physical range between the ambient

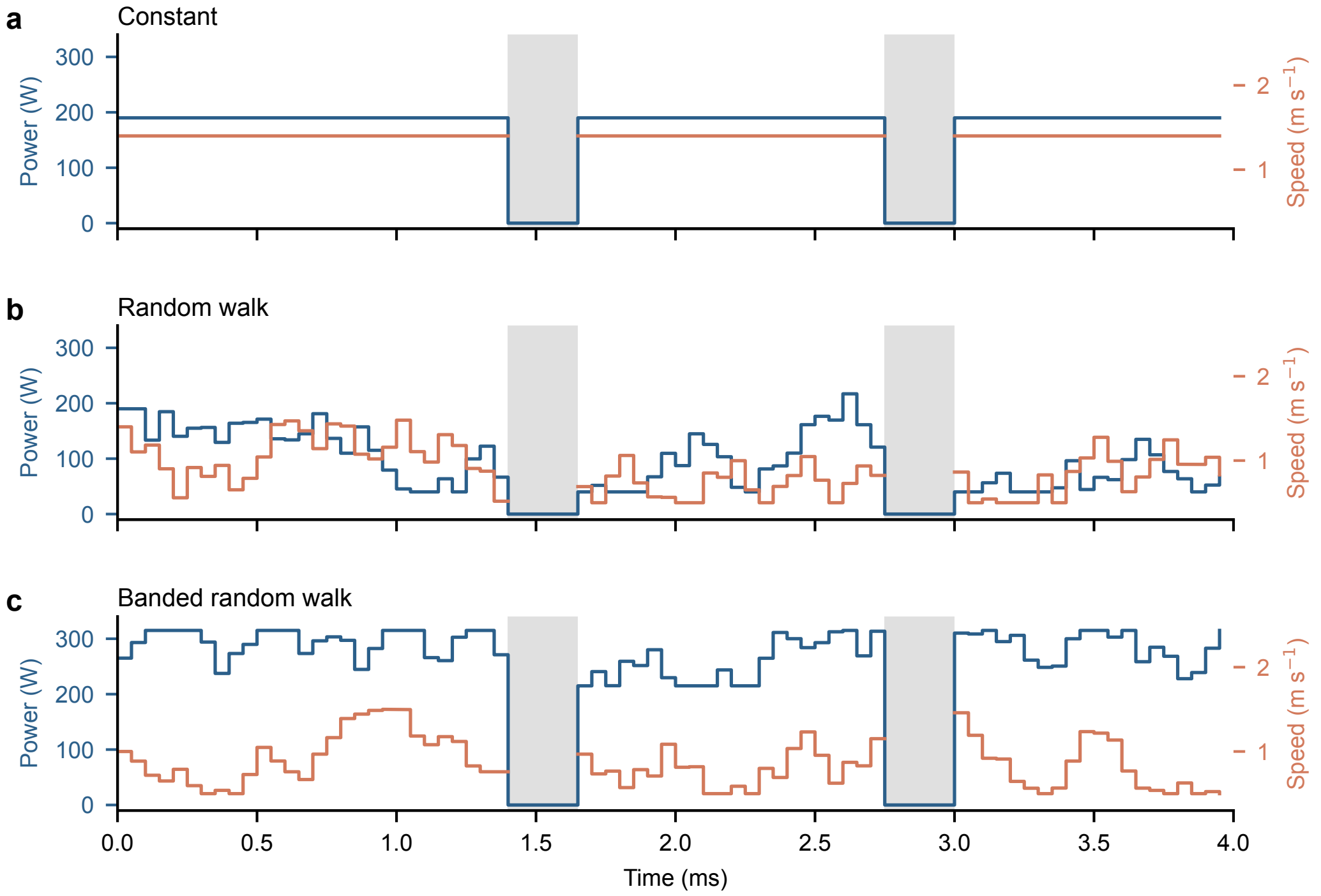


Fig. S2: **Example laser schedules of the three dataset types.** Laser power (blue, left axis) and scan speed (orange, right axis) over one 4 ms case for **(a)** constant parameters, **(b)** the clipped random walk, and **(c)** the banded random walk. Gray bands mark 0.25 ms repositioning intervals between tracks, during which the laser is off. The track boundaries shown are illustrative, as actual track lengths depend on the sampled scan path. Traces are generated with the sampling rules of Section S2.1.

Table S3: Composition of the simulated datasets. Each case uses an independently sampled substrate geometry, surface morphology, initial temperature field and scan path (Section S2.1). The schedule and banded-schedule datasets form one 800-case pool, from which the fine-tuning set ($50 + 50$ cases) and the shared validation set (40 cases) are drawn. The Pure Transient reference is trained on all 800 cases. The control-benchmark cases take their initial conditions from the rollout and transient test sets.

| Dataset | Cases | Laser schedule | Overhang angle | Role |
|---|---|---|---|---|
| Pretraining | 400 | constant ($20 \times 20$ grid) | $\mathcal{U}(10°, 60°)$ | AE + transition pretraining |
| Schedule | 400 | random walk | $\mathcal{U}(10°, 60°)$ | AE fine-tuning, time-varying pool |
| Banded schedule | 400 | banded random walk | $\mathcal{U}(45°, 60°)$ | time-varying pool |
| Rollout test | 25 | constant (off-grid) | $\mathcal{U}(10°, 60°)$ | base model evaluation |
| Transient test | 88 | step, ramp, modulated, fast walk | $\mathcal{U}(10°, 60°)$ | transition evaluation and ablation |
| Control benchmark | 30 | controller-defined | $\mathcal{U}(10°, 60°)$ | control evaluation |

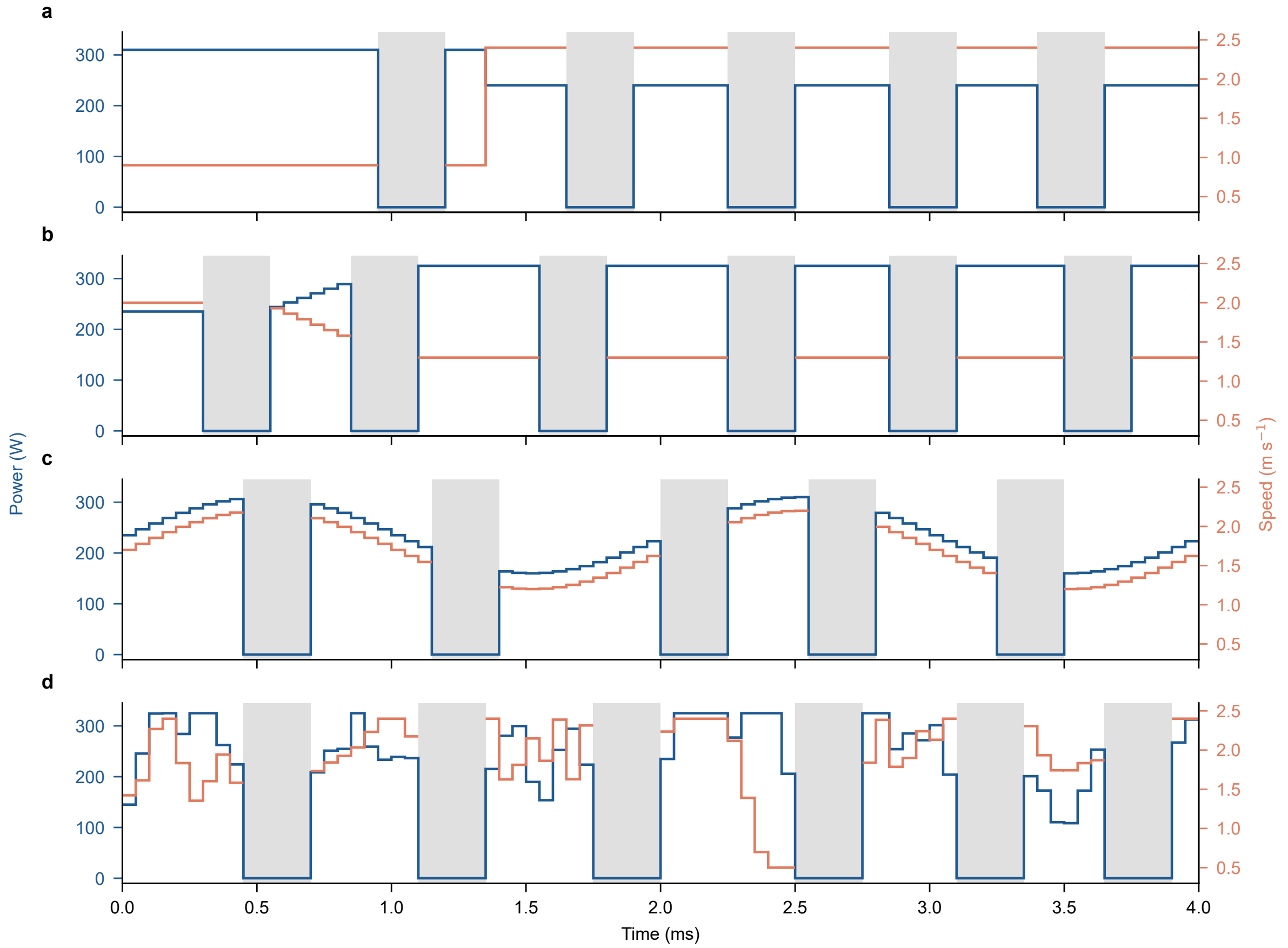


Fig. S3: **Example laser schedules of the four transient test regimes.** Laser power (blue, left axis) and scan speed (orange, right axis) recorded over one 4 ms case for **(a)** a step change in both power ($-70$ W) and speed ($+1.5$ m s$^{-1}$), **(b)** a 0.5 ms ramp in both power ($+90$ W) and speed ($-0.7$ m s$^{-1}$) starting at 0.5 ms, **(c)** a 0.5 kHz sinusoidal modulation of power (75 W amplitude) and speed (0.5 m s$^{-1}$ amplitude), and **(d)** a fast random walk with per-step changes of up to 120 W and 0.8 m s$^{-1}$. Gray bands mark the 0.25 ms repositioning intervals during which the laser is off. Each control value is held for one 50 $\mu$s control interval.

temperature and 4000 K. The phase-indicator and liquid-fraction fields lie in $[0, 1]$ and are used unnormalized.

For transition-model training, all frames of a dataset are encoded once by the frozen autoencoder into latent tensors of shape $10 \times 25 \times 25 \times 8$. Training samples are windows of $h+1 = 5$ consecutive latent frames ($h = 4$) together with up to five rollout targets. A 4 ms case yields 76 windows. The conditioning vector of each transition is $\mathbf{c}_t = (x_t^L, y_t^L, P_t)$, comprising the in-plane heat-source coordinates at the start of the interval in millimetres and the laser power applied during it in kilowatts. Gaussian noise with standard deviation $10^{-4}$ is added to the past latent window at load time as a stabilizing perturbation. The autoencoder uses a random 90/10 train-validation split over frames, drawn with a fixed seed (42). The transition models use case-level splits, so that no window of a validation case enters training: the base model and the two deterministic baselines are trained on 360 cases of the pretraining dataset and validated on the remaining 40, and each fine-tuned

model is trained on its $n$ cases and validated on a shared set of 40 cases (20 random-walk and 20 banded) from the 800-case time-varying pool, disjoint from every fine-tuning set. The fine-tuning sets are nested ($n = 10 \subset 20 \subset 50 \subset 75 \subset 100$) and balanced between random-walk and banded cases. Only the Pure Transient reference uses a random 90/10 split, over the latent windows of its 800 training cases (seed 42). The case lists are released with the code. Because early frames are underrepresented relative to their importance for rollout initialization, frames at $t = 0$ and $t = 1$ are oversampled with replacement at 15× and 7.5× their base probability.

# S4 Autoencoder architecture and training

## S4.1 Architecture

The branched FSQ autoencoder maps the two observed fields to a 10-channel latent tensor at 4× reduced resolution and reconstructs three fields (Table S4). The encoder is a 3D convolutional network with two downsampling stages, channel widths 64/128/256, two residual blocks per resolution and two mid-level residual blocks, using GroupNorm (32 groups) and SiLU activations. Its 10 output channels are split into 8 morphology channels, quantized by finite scalar quantization [10] with nominal per-channel levels $[5, 4, 4, 3, 3, 3, 2, 2]$, and two amplitude channels, one per field, each quantized uniformly on $[-1, 1]$. Each channel is quantized as $\mathrm{round}(\tanh(\cdot)\,(L-1)/2)$ without the half-step offset that FSQ applies to an even number of levels $L$, so a channel with $L = 4$ takes three values and a channel with $L = 2$ takes only the value zero. The effective morphology levels are therefore $[5, 3, 3, 3, 3, 3, 1, 1]$: six active channels with an implicit vocabulary of 1215 spatial patterns per latent site. The amplitude channels take 1023 uniformly spaced levels. Two lightweight decoder branches with mirrored architecture (base width 32, one residual block per resolution) reconstruct temperature and phase indicator from the shared morphology channels and the respective amplitude channel. A Softplus output activation enforces non-negative temperature and a Sigmoid keeps the phase indicator in $[0, 1]$. The full model has 25.6 M parameters (15.9 M encoder, 4.8 M per decoder branch).

The liquid-metal fraction is not learned as an independent output. It is computed deterministically from the reconstructed fields by the melt-fraction map $\mathcal{M}$ of the main text, applied voxel-wise to the denormalized temperature,

$$\hat{\boldsymbol{\mu}} = \mathcal{M}\left(\hat{\mathbf{T}}, \hat{\boldsymbol{\alpha}}\right) = \hat{\boldsymbol{\alpha}} \cdot \frac{1}{2}\left[1 + \tanh\left(\frac{4\left(\hat{\mathbf{T}} - T_{\mathrm{mid}}\right)}{\Delta T}\right)\right], \qquad T_{\mathrm{mid}} = 1900\ \mathrm{K}, \quad \Delta T = 46\ \mathrm{K}, \tag{S5}$$

where $T_{\mathrm{mid}}$ and $\Delta T$ are the midpoint and width of the melting interval. The phase-indicator factor restricts $\hat{\boldsymbol{\mu}}$ to cells containing material, and the smooth step transitions from 0 to 1 as the temperature crosses the melting interval, mirroring the melt-fraction definition of the CFD solver. The same map produces the decoded liquid fraction used by the auxiliary supervision of the transition model (Section S5.2) and by the depth extraction of the controllers.

## S4.2 Training

The autoencoder is trained on individual frames in two stages (Table S5): it is first trained from random initialization on the constant-schedule pretraining dataset (400 cases), and then fine-tuned, starting from these weights, on the random-walk schedule dataset (400 cases) to adapt the representation to time-varying laser schedules. The loss is the sum of (i) equally weighted mean-squared

Table S4: Autoencoder architecture.

| Component | Setting |
|---|---|
| Input | $2 \times 100 \times 100 \times 32$ (normalized $\mathbf{T}$, $\boldsymbol{\alpha}$) |
| Latent | $10 \times 25 \times 25 \times 8$ |
| Latent split | 8 morphology + 1 $T$-amplitude + 1 $\alpha$-amplitude channels |
| FSQ levels (morphology) | nominal $[5, 4, 4, 3, 3, 3, 2, 2]$ |
| Effective levels | $[5, 3, 3, 3, 3, 3, 1, 1]$ (1215 patterns per site) |
| Amplitude quantization | 1023 uniform levels per channel |
| Encoder | base width 64, multipliers $(1, 2, 4)$, 2 residual blocks per level |
| Decoders ($\times 2$) | base width 32, multipliers $(1, 2, 4)$, 1 residual block per level |
| Normalization / activation | GroupNorm (32 groups) / SiLU |
| Output activations | Softplus ($T$), Sigmoid ($\alpha$) |
| Parameters | 25.6 M |

reconstruction errors of the three output fields, (ii) a phase-mass term (weight 1) penalizing the squared relative error of the reconstructed phase indicator, (iii) a temperature-gradient term (weight 1) matching three-dimensional finite-difference gradients of the reconstructed and target temperature fields, (iv) an initial-condition penalty on $t = 0$ frames (value and gradient errors, with weight 1 during the first epoch and $10^{-6}$ thereafter), and (v) the quantizer commitment terms of the FSQ and amplitude tokenizers. Optimization uses AdamW with weight decay excluded for bias parameters, a cosine learning-rate schedule with warmup over the first 10% of steps, and bf16 mixed precision with data parallelism over four GPUs. The checkpoint with the lowest validation relative $L_2$ error is retained, and it reaches a validation relative $L_2$ error of about 1.6%. The two stages serve different purposes. The pretrained checkpoint is the FSQ arm of the representation ablation (Section S6.1 and main-text Fig. 3a,b), compared with a continuous bottleneck trained on the same constant-schedule frames. The fine-tuned checkpoint is the world-model autoencoder: it is frozen and encodes the latents on which every transition model, including the deterministic baselines, is trained.

Table S5: Autoencoder training protocol.

| | Pretraining | Fine-tuning |
|---|---|---|
| Dataset | constant schedules (400 cases) | random walk (400 cases) |
| Initialization | random | pretrained weights |
| Optimizer | AdamW | AdamW |
| Peak learning rate | $2 \times 10^{-4}$, then $5 \times 10^{-5}$ | $5 \times 10^{-5}$ |
| Weight decay | $5 \times 10^{-5}$, then $10^{-5}$ | $10^{-5}$ |
| Schedule | cosine, 10% warmup | cosine, 10% warmup |
| Batch size (global) | 8 | 8 |
| Epochs | 294 $(193 + 101)$ | checkpoint at epoch 78 of 101 |
| Precision | bf16 | bf16 |
| Hardware | 4× RTX 4090 | 4× RTX 4090 |
| Wall-clock time | $\approx 39$ h | $\approx 15$ h |

# S5 Transition model architecture and training

## S5.1 Architecture

The transition model is a diffusion transformer [11] operating on the latent grid (Table S6). The noisy residual is concatenated channel-wise with the past window of $h + 1 = 5$ latent frames (50 channels), giving 60 input channels, and divided into anisotropic $2 \times 2 \times 1$ patches (after replicate-padding the $25 \times 25$ in-plane extent to $26 \times 26$), producing $13 \times 13 \times 8 = 1352$ tokens. The depth direction retains its full latent resolution because melt-pool penetration depth is the principal controlled quantity. Tokens are embedded to width 256 with a learned positional embedding and processed by 8 transformer blocks with 8 attention heads and GELU MLPs of expansion 4. The conditioning vector $\mathbf{c}_t$ and the flow-matching time $\tau$ are embedded and injected through adaLN-Zero modulation of every attention and MLP block, with the modulation and output projections initialized to zero so that the network starts as an identity map. The model has 10.2 M parameters.

At rollout time, the flow is integrated with four Euler steps from a Gaussian sample, so that each transition costs four network evaluations. The predicted residual is added to $\mathbf{z}_t^{\mathrm{q}}$, and the result can optionally be re-quantized by the frozen FSQ tokenizer. Fixing the noise seed makes the sampled transition deterministic, and the rollout is differentiable end to end, with gradients passed through the optional FSQ rounding by the straight-through estimator, which is exploited by both controllers.

Table S6: Transition-model (DiT) architecture.

| Component | Setting |
|---|---|
| Input | noisy residual $\oplus$ past window: $60 \times 25 \times 25 \times 8$ |
| Patch size | $2 \times 2 \times 1$ (1352 tokens) |
| Width / depth / heads | 256 / 8 / 8 |
| MLP expansion | 4 (GELU) |
| Conditioning | $\mathbf{c}_t = (x_t^L, y_t^L, P_t)$ and $\tau$ via adaLN-Zero |
| Euler steps per transition | 4 |
| Parameters | 10.2 M |

## S5.2 Training

The flow-matching transition is trained in two stages with the objective of the main text (Table S7): first from random initialization on the latents of the pretraining dataset (the base model, used for the rollout-fidelity evaluation of main-text Fig. 2 and as the base of the transition ablation, trained on an explicit split of 360 training and 40 validation cases), and then initialized from the pretraining-data model and fine-tuned on 100 time-varying cases (50 schedule and 50 banded-schedule cases from the 800-case pool of Section S2.2), with early stopping on a separate 40-case validation set from the same pool (used for the control studies). The same fine-tuning recipe with $n = 10$, 20, 50 and 75 cases, drawn as nested subsets of the same case pool, provides the data-scaling study of main-text Fig. 3g,l (training curves in Fig. S4b,c). The Pure Transient reference of main-text Fig. 3 uses the architecture and training recipe of the base model (Table S7, first column) but is trained from random initialization on all 800 cases of the time-varying pool, including the fine-tuning and validation cases, with a random 90/10 split of their latent windows (seed 42) and early stopping on

this validation error. Optimization uses AdamW with a 500-step linear warmup followed by cosine decay, gradient-norm clipping at 1.0, and bf16 mixed precision over four GPUs. The checkpoint with the lowest validation error is retained, where validation draws a full four-step sample of the next latent state and measures its relative $L_2$ error (Fig. S4a,b). Both models are trained in the latent space of the frozen autoencoder.

Three mechanisms stabilize long autoregressive rollouts. First, a push-forward curriculum is applied: after a short teacher-forced warmup (3 epochs for both runs), each batch draws a rollout depth $R \sim \mathcal{U}\{1, \ldots, 5\}$, advances the first $R-1$ steps with the model's own four-step samples without gradient tracking, and applies the flow-matching loss at step $R$. Second, each sample's loss is weighted by $1 + 39\exp(-t/0.6)$, where $t$ is the frame index, emphasizing the early transient during which the melt pool forms. Third, for both models, an auxiliary supervision anchors the latent dynamics to the physical melt pool: the one-step estimate $\hat{\mathbf{x}}_1 = \mathbf{x}_\tau + (1-\tau)\mathbf{v}_\theta$ is decoded through the frozen decoder and its liquid-metal fraction is penalized with an $L_1$ loss against the CFD reference (weight $10^{-2}$).

Table S7: Transition-model training protocol.

| | Pretraining-data model | Control model |
|---|---|---|
| Training latents | pretraining ($360 + 40$ cases) | schedule + banded ($50 + 50$ cases) |
| Initialization | random | pretraining-data model |
| Latent windows | 27,360 | 7,600 |
| Optimizer | AdamW | AdamW |
| Peak learning rate | $10^{-4}$ | $5 \times 10^{-5}$ |
| Weight decay | $10^{-4}$ | $10^{-4}$ |
| Warmup steps | 500 | 200 |
| Gradient clipping | 1.0 | 1.0 |
| Batch size (global) | 32 | 32 |
| Push-forward warmup epochs | 3 | 3 |
| Liquid-fraction supervision | $L_1$, weight $10^{-2}$ | $L_1$, weight $10^{-2}$ |
| Stopping | best validation, 400 epochs (342,000 steps) | early stopping (patience 20), $\leq 60{,}000$ steps |
| Precision | bf16 | bf16 |
| Hardware | 4× RTX 4090 | 4× RTX 4090 |
| Wall-clock time | $\approx 51$ h | $\approx 9$ h |

# S6 Baselines and evaluation protocol

## S6.1 Baseline models

The transition ablation compares the flow-matching DiT against two deterministic regressors trained on the same frozen latent space, with the same past window, conditioning construction and early-step weighting. The ViT baseline [12] shares the DiT's tokenization and transformer dimensions: identical patch size ($2 \times 2 \times 1$), width (256), depth (8) and head count (8), predicting the latent

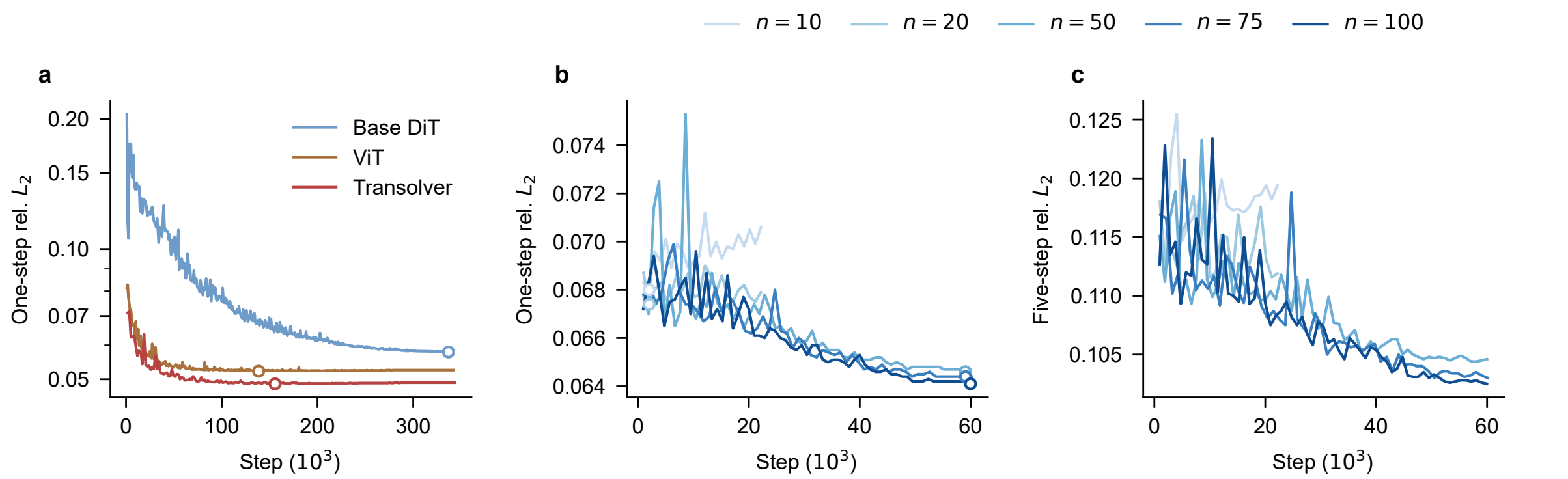


Fig. S4: **Training curves of the transition models.** **(a)** Validation one-step relative $L_2$ error of the transition models trained on the constant-schedule latents with the same 360/40 case split: the base flow-matching model (DiT) and the deterministic ViT and Transolver baselines, plotted against optimizer steps at a matched budget of approximately $3.4 \times 10^5$ steps. For the DiT the one-step prediction is one sample of the predicted distribution, whereas the deterministic regressors predict its conditional mean, which minimizes the expected squared error by construction. Their lower one-step error therefore does not carry over to recursive rollouts (main-text Fig. 3c-f). **(b)** One-step and **(c)** five-step-rollout validation relative $L_2$ error of the fine-tuning ladder, in which the base model is fine-tuned on $n = 10$, 20, 50, 75 and 100 time-varying cases and validated on the same 40 time-varying cases. With $n = 10$ and 20 the validation error does not improve beyond its value after 2,016 steps, and early stopping ends both runs after 22,176 steps. With $n \geq 50$, both errors decrease throughout, and the lowest validation error falls near the 60,000-step cap. Open circles mark the checkpoints with the lowest validation error, which are the models used in the main text. Curves are parsed from the training logs.

residual deterministically from the 50-channel past window under a mean-squared-error loss (6.8 M parameters, compared with 10.2 M for the DiT). The Transolver baseline [13] uses five physics-attention layers of width 256 with 8 heads and 64 slices, with the conditioning injected through feature-wise modulation. Both baselines are trained with AdamW (learning rate $10^{-4}$, weight decay $10^{-4}$, 500 warmup steps, gradient clipping 1.0) with a rollout curriculum of depth $R \sim \mathcal{U}\{1, \ldots, 5\}$. Because their transitions are deterministic, gradients are propagated through the full rollout with activation checkpointing rather than by push-forward. The ViT is trained for 400 epochs at a global batch size of 32 (342,000 optimizer steps), identical to the base flow-matching model, and the Transolver (19.0 M parameters) for 201 epochs at a global batch size of 16 (343,710 steps). All three constant-schedule models are thus trained with the same budget of approximately $3.4 \times 10^5$ optimizer steps on the same 360/40 split. The Transolver validation error reaches its minimum after $1.6 \times 10^5$ steps and remains flat thereafter (Fig. S4a), so its training is converged within this budget.

The representation comparison contrasts the FSQ autoencoder with a continuous bottleneck of identical architecture, in which only the rounding of the morphology and amplitude channels is removed while their tanh bounding is retained, so both models have the same parameters and latent dimensionality (25.6 M parameters, $10 \times 25 \times 25 \times 8$ latent). The FSQ arm is the autoencoder after constant-schedule pretraining (Table S5), before fine-tuning on random-walk frames. The continuous arm is trained from random initialization on the same constant-schedule dataset with

the pretraining recipe of Table S5 over a 200-epoch cosine schedule, and its epoch-150 checkpoint is used, so it receives less training than the FSQ arm. Both are evaluated on the 25 constant-parameter and 88 transient test cases by the per-frame relative $L_2$ error of each field, reported as the median over frames (for $\boldsymbol{\mu}$, over frames containing melt). The error budget encodes and decodes each CFD frame with the world-model autoencoder, without the transition model, and scores the reconstruction with the same melt-pool depth and field metrics as the rollouts (Table S8).

Table S8: Reconstruction errors of the FSQ and continuous autoencoders and error budget of the world model. Field errors are median per-frame relative $L_2$ errors in percent. The depth error is the mean absolute error of the reconstructed melt-pool depth for the autoencoders (all 80 steps for the constant set, laser-on steps for the transient set) and the depth CRPS for the world-model rollout. The world-model autoencoder is the fine-tuned FSQ autoencoder used by all transition models, and the rollout is the base transition model of main-text Figs. 2 and 3.

| Test set | Model | $\mathbf{T}$ (%) | $\boldsymbol{\alpha}$ (%) | $\boldsymbol{\mu}$ (%) | Depth ($\mu$m) |
|---|---|---|---|---|---|
| Constant (25) | FSQ autoencoder | 1.73 | 2.45 | 6.9 | 0.23 |
| | Continuous autoencoder | 1.33 | 1.86 | 5.3 | 0.14 |
| | World-model autoencoder | 1.65 | 2.65 | 6.1 | 0.17 |
| | World-model rollout | 7.49 | 7.60 | – | 4.55 |
| Transient (88) | FSQ autoencoder | 1.80 | 2.44 | 7.4 | 0.22 |
| | Continuous autoencoder | 1.37 | 1.79 | 5.6 | 0.15 |
| | World-model autoencoder | 1.69 | 2.66 | 6.4 | 0.18 |
| | World-model rollout | 8.99 | 7.79 | – | 5.77 |

## S6.2 Rollout evaluation

Rollout fidelity is evaluated on the 25 rollout-test cases. Each evaluation encodes the CFD state at $t = 0$, tiles it to fill the five-frame history, and rolls the transition model forward for 80 steps (4 ms) under the recorded heat-source controls, with four Euler steps per transition, repeated for 8 noise seeds per case. Fields are decoded at every step. Consistent with the CFD solver's boiling-temperature cap, decoded temperatures are clamped at 3315 K. Reported metrics are the relative $L_2$ errors of the physical temperature and phase-indicator fields, the intersection-over-union of the melt-pool region ($\boldsymbol{\mu} > 0.5$), and the melt-pool depth, measured from the initial substrate top surface to the deepest point at which the decoded liquid fraction reaches 0.5 and located with sub-voxel interpolation. The depth-error distribution pools all $25 \times 80 = 2000$ per-step predictions, with the predicted depth taken as the median over an ensemble of 8 rollouts with distinct noise seeds.

The distributional comparison of the transition ablation uses the continuous ranked probability score. For the flow-matching model, an ensemble of 8 rollouts with distinct noise seeds is generated per trajectory, and the CRPS against the CFD reference $y$ is computed with the empirical ensemble estimator

$$\mathrm{CRPS} = \frac{1}{M}\sum_{m=1}^{M} |x_m - y| - \frac{1}{2M^2}\sum_{m=1}^{M}\sum_{m'=1}^{M} |x_m - x_{m'}|, \tag{S6}$$

with $M = 8$ members $x_m$. For the temperature, phase-indicator and melt-pool fields it is evaluated per voxel and averaged over all voxels of the $100\times100\times32$ grid, gas cells included, whereas for melt-pool depth and volume it is evaluated per step. For the deterministic baselines ($M = 1$) the CRPS

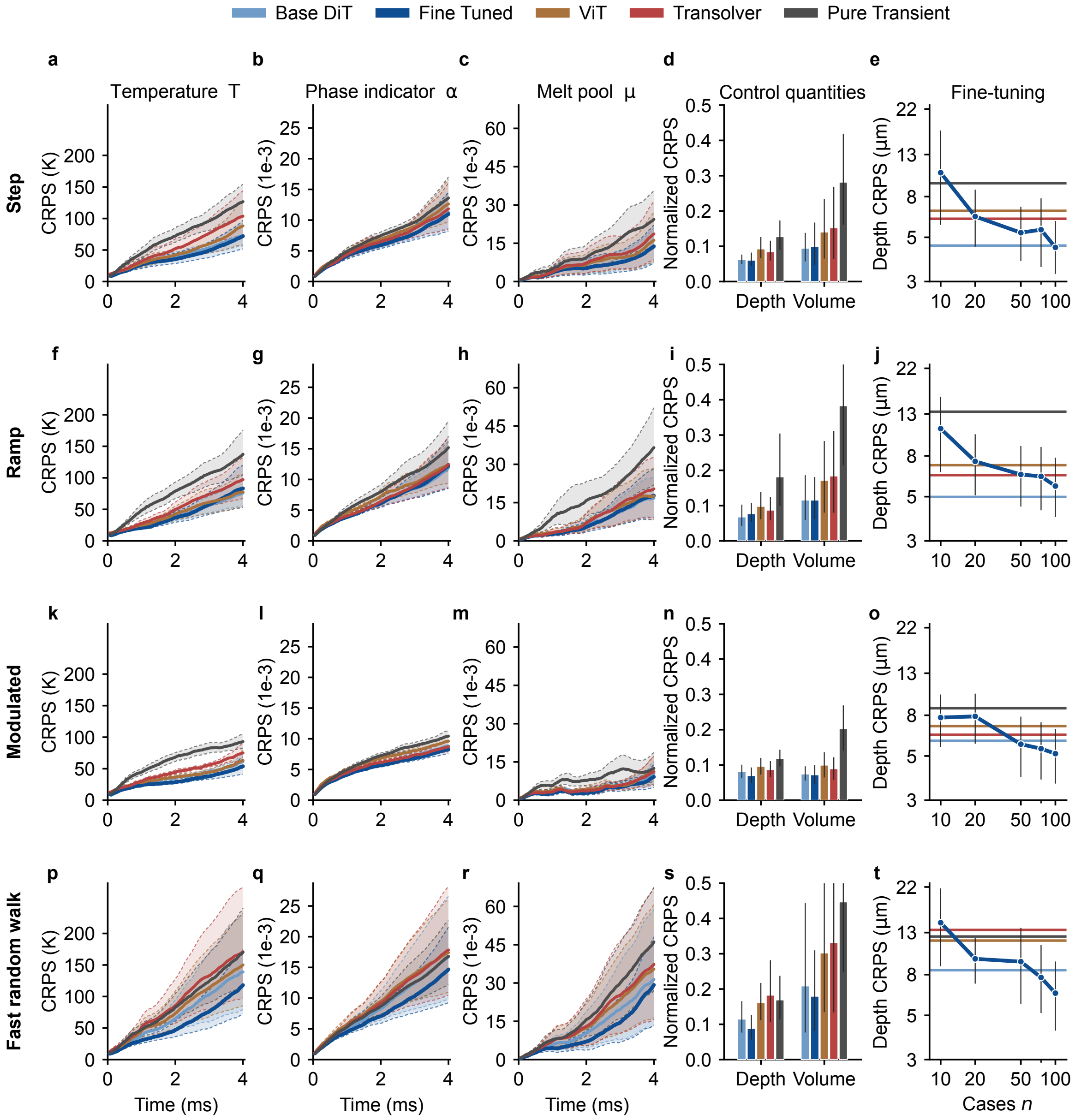


Fig. S5: **Transition ablation resolved by transient schedule regime.** Rows correspond to the step (30 cases), ramp (20), modulated (19) and fast random-walk (19) regimes of the transient test set (Fig. S3), and columns follow main-text Fig. 3c-g, with the same models, normalization and axis limits. **(a,f,k,p)** Temperature, **(b,g,l,q)** phase-indicator and **(c,h,m,r)** melt-pool-field CRPS over the 4 ms rollout. Lines are case means and shaded bands their 95% bootstrap confidence intervals. **(d,i,n,s)** Normalized CRPS of melt-pool depth and volume, with 95% bootstrap confidence intervals. **(e,j,o,t)** Melt-pool depth CRPS of the fine-tuned model against the number of time-varying fine-tuning cases $n$, with horizontal lines marking the models without fine-tuning. Paired statistics are given in Section S6.2.

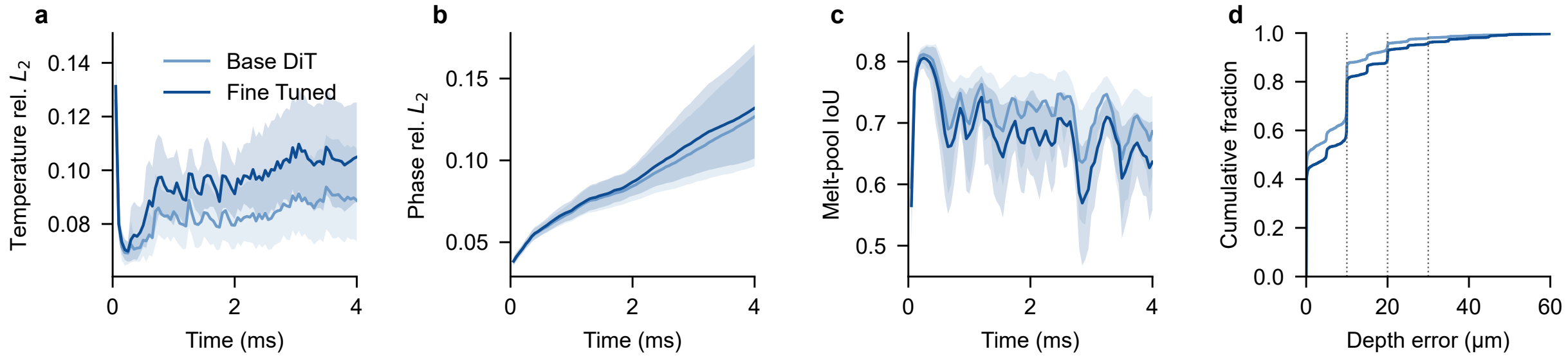


Fig. S6: **Base and fine-tuned transition on the constant-parameter test set.** Rollouts on its 25 cases of main-text Fig. 2. **(a)** Temperature and **(b)** phase-indicator relative $L_2$ error and **(c)** melt-pool intersection over union over the 4 ms rollout. Lines are case means (each case averaged over 8 noise seeds) and bands their 95% bootstrap confidence intervals. **(d)** Cumulative distribution of the melt-pool depth error, with the depth taken as the median over 8 noise seeds and all $25 \times 80 = 2000$ per-step predictions pooled, with dotted lines marking one, two and three voxels. Within one, two and three voxels lie 81%, 95% and 98% of the base-model predictions and 74%, 91% and 96% of the fine-tuned predictions.

reduces to the mean absolute error, allowing a direct comparison. These ensemble evaluations use the 88 transient test trajectories (Section S2.1) of 80 steps each, and results are reported both for all 88 cases and resolved by schedule regime (Fig. S5). Melt-pool depth and volume are scored only on laser-on steps, excluding the 0.25 ms repositioning intervals. Their CRPS is pooled over all laser-on steps of a case set, so that each step carries equal weight (4,616 steps for the 88 transient cases and 972 for the 19 fast random-walk cases). The normalized values of main-text Fig. 3f,k divide these pooled means, in both rows, by the mean CFD value over the laser-on steps of all 88 cases (73.5 $\mu$m for depth and $9.20 \times 10^{-3}$ mm$^3$ for volume), and the fine-tuning curves of main-text Fig. 3g,l show the same pooled depth CRPS in micrometres. Error bands and error bars are 95% confidence intervals from a bootstrap over cases (20,000 resamples), in which each resample draws whole cases and recomputes the step-pooled mean. Differences between models are instead computed per case: each case's value is its mean over its laser-on steps, the paired differences are averaged over cases with equal weight, and a paired bootstrap over the identical case set (20,000 resamples) gives a two-sided $P$ value. Pooled levels and case-averaged differences can therefore differ slightly. On the fast random walk, for example, the pooled base and fine-tuned depth CRPS are 8.42 and 6.46 $\mu$m, whereas the mean per-case difference is $-2.03$ $\mu$m ($P = 0.29$).

Resolving the transition ablation by schedule regime (Fig. S5) shows a consistent ordering of difficulty: every model is most accurate on step inputs and least accurate on the fast random walk, with base-model depth CRPS of 4.5, 5.0, 6.0 and 8.4 $\mu$m for step, ramp, modulated and fast random-walk inputs. The benefit of fine-tuning follows the non-stationarity of the input. On step and ramp inputs, where the melt pool relaxes toward a quasi-steady state between changes, the fine-tuned model is not more accurate than the base model ($-0.07$ $\mu$m, $P = 0.84$, and $+0.69$ $\mu$m, $P = 0.15$, respectively). On modulated inputs it improves the depth CRPS by 0.90 $\mu$m ($P = 0.020$), and on the fast random walk it lowers it from 8.4 to 6.5 $\mu$m, the largest change in magnitude but not significant with 19 cases ($P = 0.29$). The degradation from fine-tuning on few cases is significant for $n = 10$ on step, ramp and modulated inputs, and persists up to $n = 75$ on ramp inputs. The flow-matching base model is more accurate than both deterministic baselines in every regime,

significantly so on step and fast random-walk inputs, and the phase indicator is predicted similarly by all models throughout.

### S6.3 Qualitative rollouts under fast random-walk inputs

Figures S7-S10 complement the aggregate transient evaluation with three-dimensional temperature and surface-geometry comparisons between the fine-tuned world model and CFD on four of the 19 fast random-walk test cases. The cases were selected by quantiles of per-case laser-on depth-CRPS differences: case p83 is the median of the fine-tuned-minus-base DiT difference, and cases p89, p88 and p84 correspond to the 25th, 50th and 75th percentiles of the ViT-minus-base DiT difference, respectively. Selection therefore uses the quantitative evaluation rather than visual agreement. Fine-tuning lowers the depth CRPS relative to the base model for p83, p89 and p84, but increases it for p88, which illustrates that the average benefit on fast random-walk inputs does not imply an improvement on every trajectory. Each figure shows one autoregressive sample (noise seed 0), whereas its header reports the ensemble-based laser-on depth CRPS used in the quantitative evaluation.

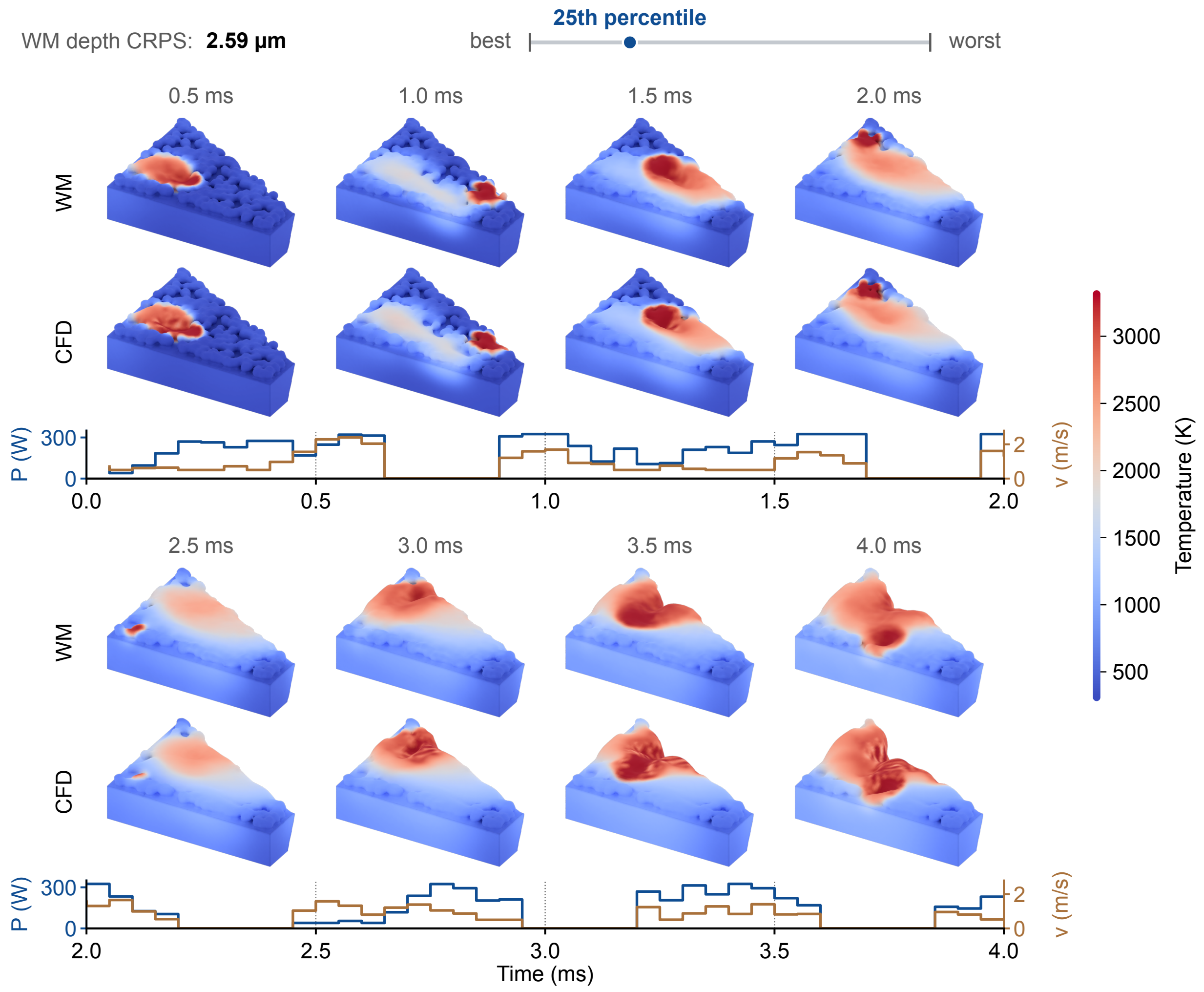


Fig. S7: **Qualitative fast random-walk rollout of the fine-tuned world model: case p83.** World-model (WM) and CFD temperature fields at eight times from 0.5 to 4.0 ms, arranged in two blocks of four columns. WM denotes the early-stopped DiT fine-tuned on $n = 100$ time-varying cases. Starting from the encoded CFD state at $t = 0$, the model advances autoregressively for 80 steps of 50 $\mu$s under the recorded controls, without subsequent CFD feedback. The displayed sample uses noise seed 0 and four function evaluations per transition. Surfaces delimit the metal region ($\alpha \geq 0.51$) and are coloured by temperature on the scale of main-text Fig. 2. Below each block, laser power $P$ (navy, left axis) and scan speed $v$ (ochre, right axis) show the inputs. The strip below each column covers the preceding 0.5 ms, and dotted lines mark the frame times. The header reports the ensemble-based laser-on depth CRPS (2.59 $\mu$m, versus 3.77 $\mu$m for the base DiT). The navy marker locates the case among the 19 fast random-walk cases from lowest to highest WM depth CRPS: rank 5 of 19, displayed as the 25th percentile after rounding to the nearest five percentage points.

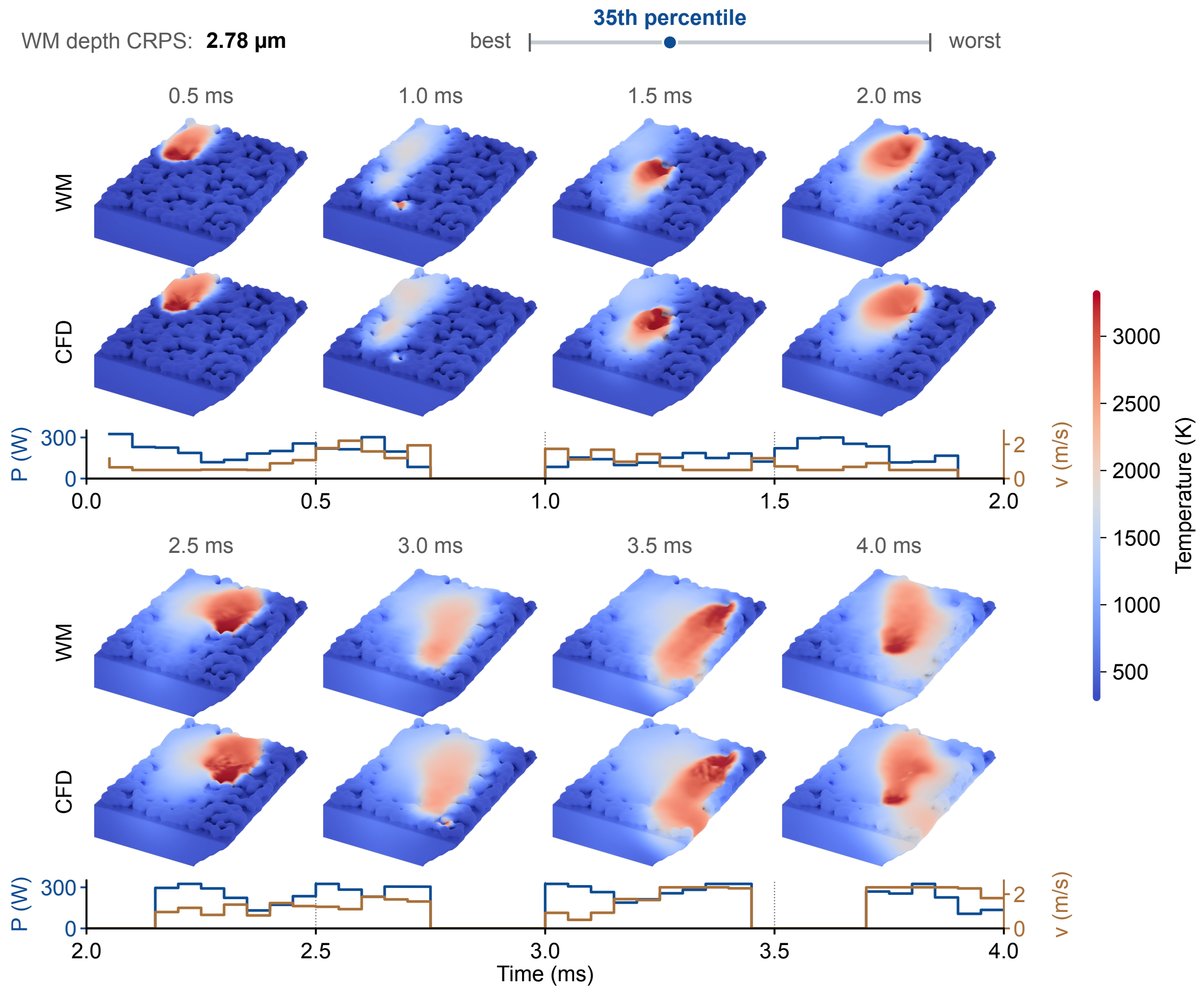


Fig. S8: **Qualitative fast random-walk rollout of the fine-tuned world model: case p89.** As in Fig. S7, for case p89, selected at the 25th percentile of the ViT-minus-base DiT depth-CRPS difference. WM and CFD temperature fields are shown at 0.5 ms intervals with the corresponding laser-power and scan-speed schedules. The displayed WM trajectory uses noise seed 0, whereas the ensemble-based laser-on depth CRPS is 2.78 $\mu$m, compared with 4.68 $\mu$m for the base DiT. This case ranks 7th of 19 fast random-walk cases by increasing fine-tuned depth CRPS, shown as the 35th percentile.

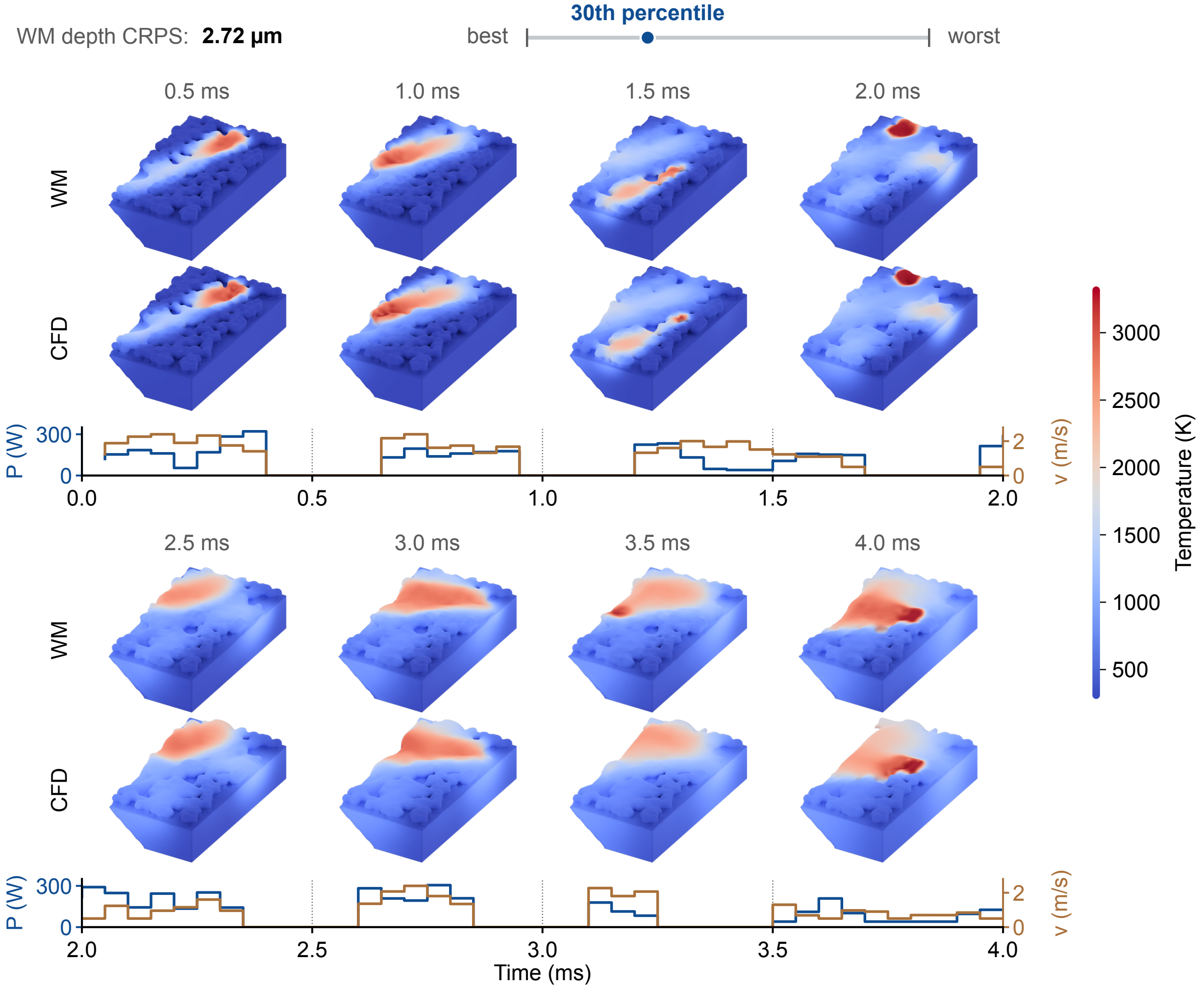


Fig. S9: **Qualitative fast random-walk rollout of the fine-tuned world model: case p84.** As in Fig. S7, for case p84, selected at the 75th percentile of the ViT-minus-base DiT depth-CRPS difference. WM and CFD temperature fields are shown at 0.5 ms intervals with the corresponding laser-power and scan-speed schedules. The displayed WM trajectory uses noise seed 0, whereas the ensemble-based laser-on depth CRPS is 2.72 $\mu$m, compared with 3.40 $\mu$m for the base DiT. This case ranks 6th of 19 fast random-walk cases by increasing fine-tuned depth CRPS, shown as the 30th percentile.

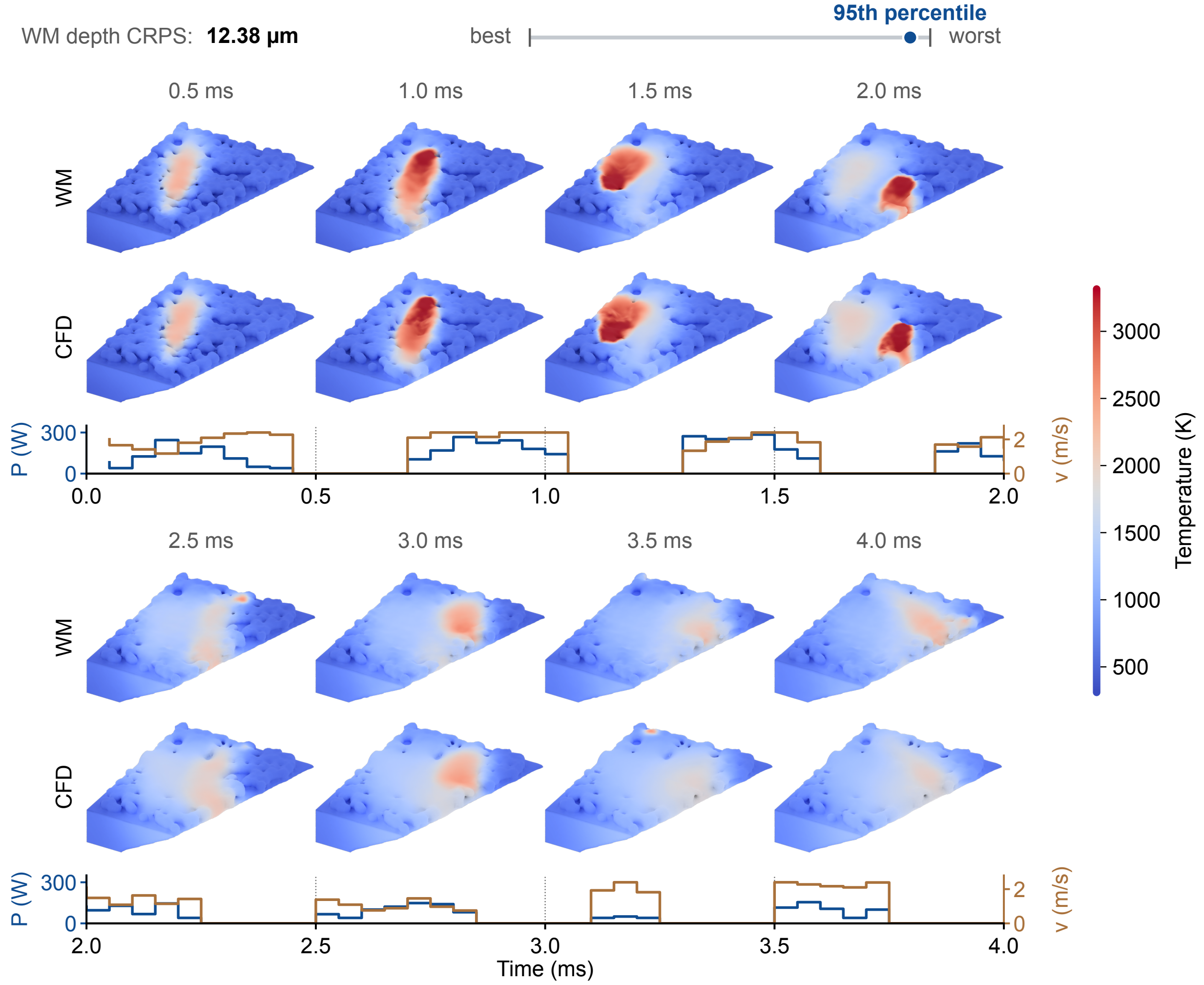


Fig. S10: **Qualitative fast random-walk rollout with degraded depth accuracy after fine-tuning: case p88.** As in Fig. S7, for case p88, selected at the median of the ViT-minus-base DiT depth-CRPS difference. WM and CFD temperature fields are shown at 0.5 ms intervals with the corresponding laser-power and scan-speed schedules. The displayed WM trajectory uses noise seed 0. The ensemble-based laser-on depth CRPS is 12.38 $\mu$m for the fine-tuned model, compared with 4.91 $\mu$m for the base DiT. This case ranks 18th of 19 fast random-walk cases by increasing fine-tuned depth CRPS, shown as the 95th percentile. This case illustrates a deterioration in depth accuracy despite the average improvement across the fast random-walk test set.

## S7 Controller configuration

### S7.1 Shared plant interface

Both controllers act on the frozen control-stage world model with the same bounded action space, $v_t \in [0.8, 2.4]$ m s$^{-1}$ and $P_t \in [80, 300]$ W, and the same control interval $\Delta t = 50$ $\mu$s. The local support thickness below the laser, $\delta_k$, is precomputed for each case from the initial substrate geometry as a $100 \times 100$ map on the simulation grid. Both controllers use the thickness-capped band penalty of Methods, with the effective depth target at step $k$

$$d_k^* = \min\bigl(d_{\text{fuse}},\ \min\bigl(\rho^* \delta_k,\ \max(d_{\text{bond}},\ \rho^* \bar{\delta}_k)\bigr)\bigr)\,, \tag{S7}$$

where $d_{\text{fuse}} = 50$ $\mu$m is the nominal fusion depth, $d_{\text{bond}} = 25$ $\mu$m a minimum bonding depth, $\rho^* = 0.5$ the relative-depth threshold, and $\bar{\delta}_k$ the support thickness at a single lookahead point 100 $\mu$m ahead of the laser along the path (one control interval at 2.0 m s$^{-1}$). The stage cost is

$$\ell_k = w_{\text{band}} \left( \frac{d_k - d_k^*}{\sigma_{\text{band}}} \right)^2 + w_{\text{var}} \left( \frac{d_k - d_{k-1}}{d_{\text{fuse}}} \right)^2 , \tag{S8}$$

with band width $\sigma_{\text{band}} = 15$ $\mu$m, $w_{\text{band}} = 5$ and $w_{\text{var}} = 1$. Here $d_k$ is a smooth relaxation of the threshold depth of Section S6.2, differentiable with respect to the decoded liquid fraction. The stage cost is summed over the laser-on steps of each planning window or imagined trajectory. Repositioning steps, during which the laser is off, contribute nothing. The target $d_k^*$ is computed from thickness lookups that are held constant during differentiation, so that the optimizer cannot lower it by changing the speed.

### S7.2 Model-predictive controller

At each replanning step, the receding-horizon controller optimizes a candidate sequence of $H = 10$ actions (0.5 ms) by backpropagating the summed stage cost through the latent rollout, and executes the first $N = 2$ actions before replanning (Table S9). The actions are parameterized by unconstrained variables $\tilde{v}_t$ and $\tilde{P}_t$ that a sigmoid maps into the bounds, $v_t = v_{\min} + (v_{\max} - v_{\min})\,\text{sigmoid}(\tilde{v}_t)$ and $P_t = P_{\min} + (P_{\max} - P_{\min})\,\text{sigmoid}(\tilde{P}_t)$, so that every candidate schedule respects the action bounds. Each replanning solves 10 Adam iterations on these unconstrained variables at learning rate 0.5 with gradient-norm clipping at 1.0. Because the scan path is fixed, the number of remaining control intervals is recomputed from the optimized speeds at each replanning, so that faster schedules finish the path in fewer intervals. To keep the optimization inexpensive, the planning rollout integrates the flow with two Euler steps per transition (rather than the four used for evaluation rollouts), with gradients propagated through the full planning window using activation checkpointing. The controls are initialized at 1.8 m s$^{-1}$ and 250 W at the start of each case and warm-started from the previous plan thereafter.

### S7.3 Amortized policy

The actor and critic share the same observation encoder (Table S10). The observation combines three inputs. The quantized latent history (50 channels at $25 \times 25 \times 8$) is processed by a three-layer 3D convolutional encoder (widths 64/64/128, GELU) followed by a linear projection to 256 features. Twelve scalars, comprising the laser position (mm), the local path tangent, the previous

Table S9: Model-predictive controller settings.

| Setting | Value |
|---|---|
| Planning horizon $H$ | 10 steps (0.5 ms) |
| Executed actions per replan $N$ | 2 |
| Action parameterization | sigmoid of unconstrained variables |
| Inner optimizer | Adam, 10 iterations, learning rate 0.5 |
| Gradient clipping | 1.0 |
| Euler steps per transition (planning) | 2 |
| Action bounds | $v \in [0.8, 2.4]$ m s$^{-1}$, $P \in [80, 300]$ W |
| Initialization | $v = 1.8$ m s$^{-1}$, $P = 250$ W |

normalized action, the scan progress, and the support thickness at five lookahead points along the path, are processed by a two-layer MLP to 32 features. A $64 \times 64$ local support-thickness patch centered on the laser is processed by a two-layer 2D convolutional encoder to 32 features. Thickness inputs are scaled by 200 $\mu$m. The concatenated 320 features feed separate two-layer heads of width 256: the actor outputs two logits mapped to the bounded actions through sigmoids, and the critic outputs a scalar value.

Training follows the SHAC scheme of Methods. Each update samples $B = 6$ initial observations by drawing a random case and a random time window from the constant-schedule training set, excluding windows whose initial melt-pool depth exceeds 80 $\mu$m. The actor is rolled through the frozen world model for $L = 16$ steps with seeded transition noise, and the bootstrapped return is backpropagated through the differentiable rollout with activation checkpointing. The critic is fitted to the backward-accumulated return targets. Actor and critic use Adam with learning rates $10^{-4}$ and $3 \times 10^{-4}$, gradient-norm clipping at 5.0, and discount $\beta = 0.99$, for 700 alternating updates on a single GPU. At deployment only the actor is retained and evaluated once per control step.

Table S10: Amortized-policy (SHAC) settings.

| Setting | Value |
|---|---|
| Rollout length $L$ | 16 steps (0.8 ms) |
| Trajectories per update $B$ | 6 |
| Updates | 700 |
| Discount $\beta$ | 0.99 |
| Actor / critic optimizer | Adam, $10^{-4}$ / $3 \times 10^{-4}$ |
| Gradient clipping | 5.0 |
| Feature dimensions | latent 256 + scalars 32 + thickness patch 32 |
| Head widths | 256, 256 |
| Action bounds | $v \in [0.8, 2.4]$ m s$^{-1}$, $P \in [80, 300]$ W |
| Actor parameters | 4.8 M |

## S7.4 Control benchmark and CFD re-validation

The control benchmark executes each of the 30 benchmark cases, whose initial conditions are drawn at random from the union of the 25-case rollout test set and the 88-case transient test set

(Section S2.2), with the unoptimized schedule, MPC and the amortized policy. The unoptimized reference holds the nominal parameters $P = 250$ W and $v = 1.8\ \mathrm{m\,s^{-1}}$ constant along the path. For every controlled trajectory, the selected power and speed schedules are exported together with the case's initial temperature and phase fields, imposed as piecewise-constant inputs over successive 50 $\mu$s intervals, and re-simulated with the identical CFD configuration of Section S1. World-model and CFD depths are then compared on the simulation grid.

# S8 Compute environment

All CFD simulations run on the Euler cluster of ETH Zurich, each case decomposed over 64 MPI processes on a single AMD EPYC 7763 node. Re-simulating one control case required a median wall-clock time of approximately 17 minutes (interquartile range 13 to 19 minutes). The solver is containerized (Docker image converted to Apptainer) with OpenFOAM v5, swak4Foam for field initialization and the isoAdvector interface scheme. Network training uses PyTorch (v2.10) with Accelerate (v1.12) for bf16 data-parallel training on nodes with four NVIDIA RTX 4090 GPUs (24 GB), and experiment tracking uses Weights & Biases. Controller training, policy inference and all latency measurements of the main text run on a single NVIDIA RTX 5090 GPU.



\end{document}